\documentclass{egpubl}

 \JournalSubmission    
 \electronicVersion 

\usepackage{graphicx}

\PrintedOrElectronic

\usepackage{egweblnk}
\usepackage{cite}

\usepackage{balance}  
\usepackage{enumitem}

\usepackage{booktabs}
\usepackage{microtype}

\usepackage{appendix}

\title[Integrating Machine Learning into Visual Analytics]%
      {The State of the Art in \\~Integrating Machine Learning into Visual Analytics}

\author[Endert et al.]
		{A. Endert$^{1}$, W. Ribarsky$^{2}$, C. Turkay$^{3}$, W. Wong$^{4}$, I. Nabney$^{5}$, I. D\'iaz Blanco$^{6}$, F. Rossi$^{7}$
		\\
         $^1$Georgia Tech, USA\\
         $^2$University of North Carolina, Charlotte, USA\\
         $^3$City University of London, UK\\
         $^4$Middlesex University, UK\\
         $^5$Aston University, UK\\
         $^6$University of Oviedo, Spain\\
         $^7$Paris 1 Panth\'eon Sorbonne University, Paris
       }

\begin{document}

\maketitle

\begin{abstract}
Visual analytics systems combine machine learning or other analytic techniques with  interactive data visualization to promote sensemaking and analytical reasoning. It is through such techniques that people can make sense of large, complex data. While progress has been made, the tactful combination of machine learning and data visualization is still under-explored. This state-of-the-art report presents a summary of the progress that has been made by highlighting and synthesizing select research advances. Further, it presents opportunities and challenges to enhance the synergy between machine learning and visual analytics for impactful future research directions. 


\end{abstract}

\section{Introduction}

We are in a data-driven era. Increasingly more domains generate and consume data.  People have the potential to understand phenomena in more depth using new data analysis techniques. Additionally, new phenomena can be uncovered in domains where data is becoming available. Thus, making sense of data is becoming increasingly important, and this is driving the need for systems that enable people to analyze and understand data. 

However, this opportunity to discover also presents challenges. 
Reasoning about data is becoming more complicated and difficult as data scales and complexities increase.
People require powerful tools to draw valid conclusions from data, while maintaining trustworthy and interpretable results. 

We claim that visual analytics (VA) and machine learning (ML) have complementing strengths and weaknesses to address these challenges.
Visual analytics (VA) is a multi-disciplinary domain that combines data visualization with machine learning (ML) and other automated techniques to create systems that help people make sense of data~\cite{thomas2005illuminating,Kerren2008Visual,Keim2002Information,Keim2006Challenges}. Over the years, much work has been done to establish the foundations of this area, create research advances in select topics, and form a community of researchers to continue to evolve the state of the art.

Currently, VA techniques exist that make use of select ML models or algorithms. However, there are additional techniques that can apply to the broader visual data analysis process. Doing so reveals opportunities for how to couple user tasks and activities with such models. Similarly, opportunities exist to advance ML models based on the cognitive tasks invoked by interactive VA techniques. 

This state-of-the-art report briefly summarizes the advances made at the intersection of ML and VA. It describes the extent to which machine learning methods are utilized in visual analytics to date.
Further, it illuminates the opportunities within both disciplines that can drive important research directions in the future. Much of the content and inspiration for this paper originated during a Dagstuhl Seminar titled, ``Bridging Machine Learning with Information Visualization (15101)'' \cite{keim_et_al:DR:2015:5266}.

\subsection{Report organization}
This report is organized as follows. Section~\ref{sec:models} of the report discusses three categories of models: human reasoning, visual analytics and information visualization, and machine learning. The models describing the cognitive activity of sensemaking and analytical reasoning characterize the processes that humans engage in cognitively to gain understanding of data. The models and frameworks for visual analytics depict systematic descriptions of how computation and analytics can be incorporated in the systematic construction and design of visual analytic applications. Finally, the machine learning community has several models that illustrate how models are trained, used, and interactively steered. 

Section~\ref{sec:category} categorizes the integration of machine learning techniques into visual analytic systems. Section~\ref{sec:domains} discusses how such systems have been used in specific domains to solve real-world challenges. Section~\ref{subsec:SteeringDR} discusses a research direction for integrating steerable dimension reduction techniques into visual analytics. Finally, Section~\ref{sec:challenges} discusses open challenges and opportunities for ML and VA. While the current work shows how some progress has been made in bringing these two communities closer together, there are several open challenges. 

\section{Models and Frameworks}

\label{sec:models}
To ground the discussion of embedding ML techniques into VA systems for data analysis and knowledge discovery, we describe three categories of models and frameworks below. First, we discuss existing models meant to describe the cognitive stages people progress through while analyzing data. These models show the complex processes people go through to gain insight from data, which developed systems must support. Second, we discuss existing models and frameworks that describe interaction and information design of visual analytic applications. These models illustrate how data transformation and analytic computation are involved in generating the visual representations of data in tools. User interaction is critical in tuning and steering the parameters of these models. Finally, we show select ML frameworks that emphasize the importance of training data and ground truth for generating accurate and effective computational models. In addition, we describe the main techniques developed in the ML field to integrate user feedback in the training process.

\subsection{Models of Sensemaking and Knowledge Discovery}
One should emphasize that a primary purpose of data analytics is for people to understand, and gain insights into, their data~\cite{Card1999Readings,Chris2006Toward}. Thus, it is important to understand the cognitive processes of people as they reason about data. It is from such an understanding that ``human-in-the-loop'' application designs are realized. Prior work exists that provides models and design guidelines for visual analytics.

\begin{figure}
  \centering
  \includegraphics[width=0.99\columnwidth]{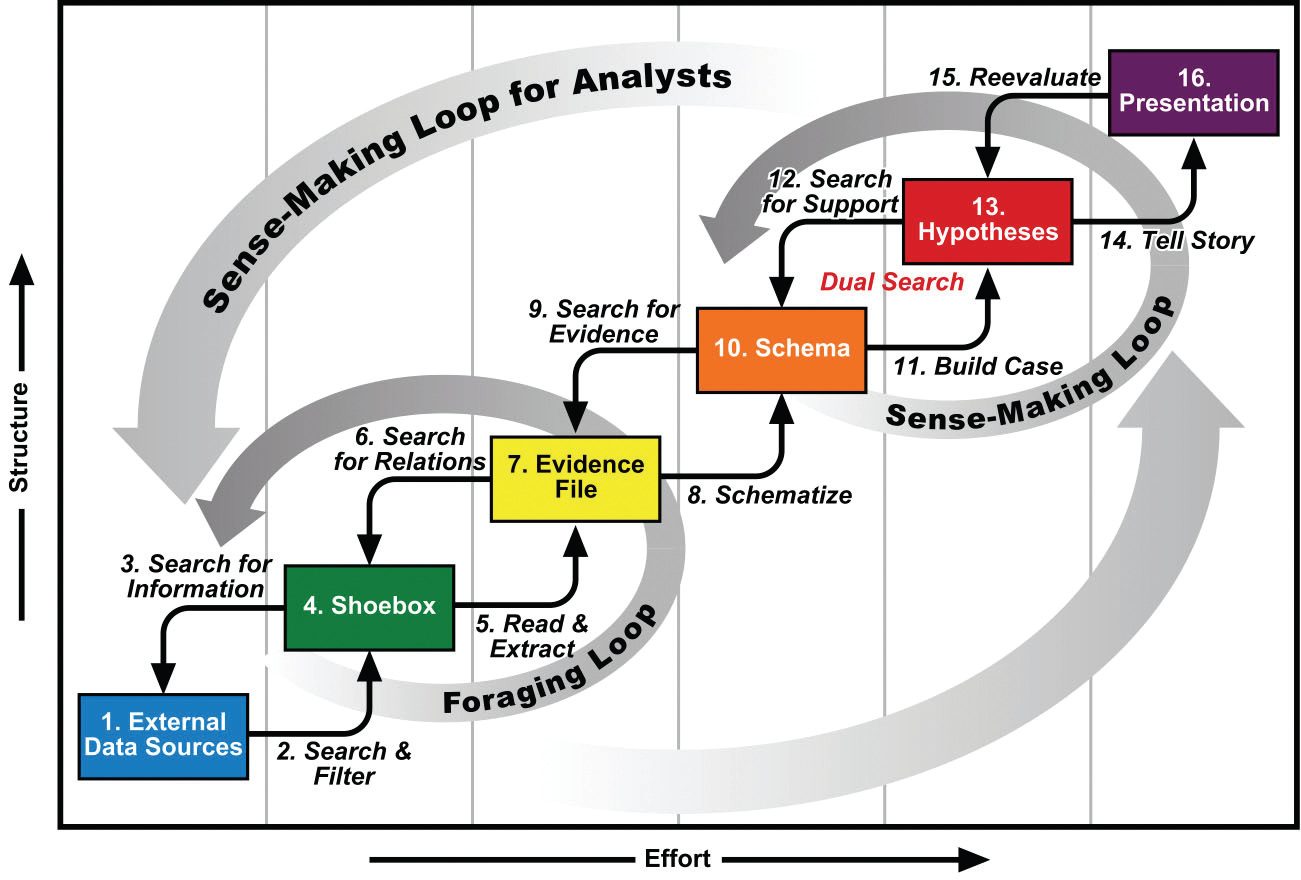}
  \caption{The ``sensemaking loop'' (from~\cite{pirolli2005sensemaking}) illustrating the cognitive stages people go through to gain insight from data. }\label{fig:pirolliCardSensemaking} 
  \vspace{-2em}
\end{figure}
Sense-making is the process of ``structuring the unknown'' by organising data into a framework that enables us ``to comprehend, understand, explain, attribute, extrapolate, and predict''~\cite{ancona2012framing}. It is this activity of structuring--the finding and assembly of data into meaningful explanatory sequences~\cite{lonergan1957study}--that enables us to turn ever more complex observations of the world into findings we can understand ``explicitly in words and that serves as a springboard into action''~\cite{weick2005organizing}. By attempting to articulate the unknown, we are driven more by ``plausibility rather than accuracy''~\cite{weick1995sensemaking} as we create plausible explanations that can be used to evolve and test our understanding of the situation or the data.  Decision makers are often faced with inaccurate representations of the world~\cite{elm2005finding} and have to fill-in the gaps with strategies such as ``story-telling'' to create stories that explain the situation.

One of the earliest models to describe the iterative process of data analysis  as ``sensemaking''~\cite{russell1993cost} is presented in Figure~\ref{fig:pirolliCardSensemaking} and illustrates the well-known (and probably the most frequently cited) Pirolli and Card sensemaking model~\cite{pirolli2005sensemaking}. Proposed in the context of intelligence analysis, it is useful for showing how information is handled through the process of searching and retrieving relevant information, organizing, indexing and storing the information for later use, structuring the information to create a schema or a way to explain what has been observed, the formulation and testing of hypotheses, which then leads to the determination of a conclusion, and a sharing of that conclusion. This notional model depicts the cognitive stages of people as they use visual analytic tools to gain understanding of their data. 

From Pirolli and Card's perspective, sensemaking can be categorized into two primary phases: foraging and synthesis. Foraging refers to the stages of the process where models filter and users gather collections of interesting or relevant information. This phase emphasizes the computational ability of models, as the datasets are typically much larger than what a user can handle. Then, using that foraged information, users advance through the synthesis stages of the process, where they construct and test hypotheses about how the foraged information may relate to the larger plot. In contrast to foraging, synthesis is more ``cognitively intensive'', as much of the insights stem from the user's intuition and domain expertise. Most existing visualization tools focus on either foraging or synthesis, separating these two phases. 

As with all models of cognitive processes, there have been criticisms. For instance, while there are feedback loops and repeat loops, and cycles within cycles, it still is somewhat a linear model. It describes the data transaction and information handling and transformation processes, ``... rather than how analysts work and how they transition''~\cite{kang2011characterizing}.  Human analysts carry out their work within this framework, but their thinking and reasoning processes are much less linear and structured. For example, although recognised as a part of the sense-making loop, there is little explanation about the thinking and reasoning strategies that are invoked to formulate hypotheses.  This is a critical aspect of the sense-making process: how are explanations of the situation or data formed in the mind of the human in order that the explanation can be used to test one's understanding of the data or situation? Later in this section, we report on work that is attempting to unravel this aspect of how analysts think.

\begin{figure}
  \centering
  \includegraphics[width=0.85\columnwidth]{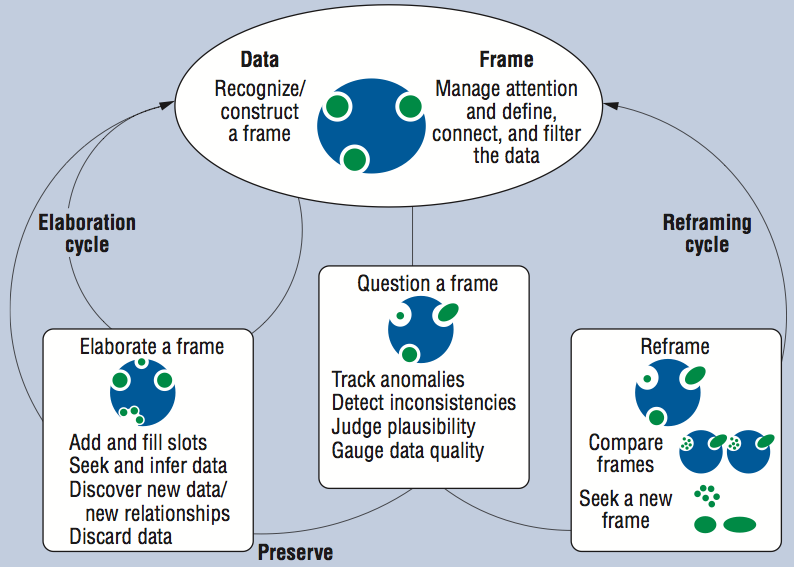}
  \caption{\label{fig:Data_Frame} The Data-Frame Model of Sense-making~\cite{Klein2006Making}.}
  \vspace{-2em}
\end{figure}

Another useful model that can be employed to describe the human-centered sense-making process is the ``data-frame model'' by Klein et al.~\cite{Klein2006Making,Klein2006Makinga}. Their model (Figure~\ref{fig:Data_Frame}) depicts an exchange of information between the human and the data in terms of frames. People make sense of a situation by interpreting the data they are presented with in relation to what they already know to create a new understanding. A user has an internal ``frame'' that represents her current understanding of the world. The data connects with the frame. As she continues to explore a particular dataset, her frames of the world are mapped against the information she uncovers. If the information supports a specific frame, that frame is thought to strengthen in a process they call elaboration. As she understands the situation better, she searches for more relevant information, learning that there may be other factors to the problem than originally thought or known, therefore driving the demand for more information, and building her frame.  However, when evidence is discovered through exploration that contradicts or refutes the existence of such a mental frame, the frame can either be augmented or a new one created. This is the important process that leads her to question her earlier conclusions or assumptions made to arrive at these conclusions. Additionally new frames can also be created to reframe the problem. In situations where data is missing or ambiguous or unknown, reframing enables her to articulate the problem in different ways that may allow her to change her information search strategy and perhaps even her goals. One of the key benefits of the Data-Frame Model is that it points to the importance of designing visual analytics in a way that encourages analysts to question their data and their understanding, and to facilitate visualizations and transformations that enable reframing of their understanding of the situation.

\begin{figure}
  \centering
  \includegraphics[width=0.99\columnwidth]{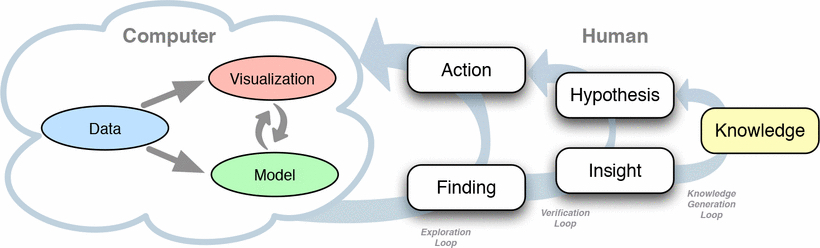}
  \caption{\label{fig:sacha_model} Human-Computer knowledge generation model of Sacha et al.~\cite{sacha2014knowledge}.}
  \vspace{-2em}
\end{figure}

Recently a set of knowledge generation and synthesis models have been proposed that comprehensively attack a central issue of visual analytics: developing a human-computer system that enables analytic reasoning to produce actionable knowledge. The first of these models was proposed by Sacha et. al.~\cite{sacha2014knowledge} and is shown in Figure~\ref{fig:sacha_model}. One sees looping structures and components familiar from Pirolli and Card's sensemaking model, as depicted in Figure~\ref{fig:pirolliCardSensemaking} above. However, the computer and human regions of the model, and their relationship with each other, are now explicitly expressed, and the paper shows a clear relationship, via interaction, between the human and both the visualization and the model. The paper also describes detailed steps for the data-visualization and data-model pipelines (the latter in terms of KDD processes that couple, for example, to machine learning algorithms). Whereas the sensemaking model was conceptual, this model is concrete and shows, better than other models, where to put computing and (via interactive interfaces) human-in-the-loop steps in order to build an actual system.

The Sacha et al. model has recently been generalized to produce a more complete knowledge generation and synthesis (KGS) model~\cite{Ribarsky2016Human}. The KGS model explicityly accounts for both Prior Knowledge (placed between Data, Visualization, and Model in Figure~\ref{fig:sacha_model}) and User Knowledge (placed between Action and Finding). Prior Knowledge is quite important for any exploration involving experts or based on expertise; experts will want to know immediately the relationship of new knowledge to existing domain knowledge. User knowledge is built up during complex reasoning, where it can then be the basis for generating additional knowledge or can be synthesized with Prior Knowledge to produce more general truths. The KGS model posits an iterative process that addresses high level reasoning, such as inductive, deductive, and abductive reasoning, in the knowledge generation and exploration loops. It is based on a framework by Gahegan et al.~\cite{gahegan2001integration} that was developed for GIScience but is generalizable. 

These models provide a roadmap for visualization and analytics processes, and for the role of human-computer interaction. In particular, they illuminate the relationships among machine learning, visualization, and analytics reasoning processes including exploration and knowledge generation. For example, Klein's data frame model, discussed above, would fit in this structure, providing a focus for ML components while the models in Figure~\ref{fig:sacha_model} would show how to connect the data frame model with interactive visualization and hypothesis-building. There are no VA systems that embody all the components of the Sacha and KGS models, but there are some (e.g., VAiRoma~\cite{cho2016vairoma}) that include parts of the model. Typically in these systems, ML is a static pre-processing step applied to the data at the beginning. For example, in VAiRoma time-dependent, hierarchical topic modeling is applied to large text collections~\cite{cho2016vairoma}. However, the KGS model shows how interactive ML can be placed in the human-computer process and how it relates to interactive visualization and reasoning. There is further discussion of interactivity in VA\/ML systems below. The discussion in Sacha et al.~\cite{sacha2014knowledge} implies two main roles for ML; one is to transform unstructured or semi-structured data into a form more meaningful for human exploration and insight discovery. The other is to use unsupervised or semi-supervised ML to guide the analysis itself by suggesting the best visualizations, sequences of steps in the exploration, verification, or knowledge generation processes, guarding against cognitive bias, etc. In addition, since the KGS model was derived with reference to cognitive science principles~\cite{Green2009Building}, there is a possibility for merging ML with cognitive models to produce even more powerful human-machine models. To illustrate, one could explore Fu and Pirolli's SNIF-ACT cognitive architecture model~\cite{fu2007snif}, which connects human exploration and information foraging in a sensemaking context. This could be married with ML approaches to refine and focus the parameters of the ML approach for particular exploration strategies.

\subsection{Models of Interactivity in Visual Analytics}

\begin{figure}
  \centering
  \includegraphics[width=0.9\columnwidth]{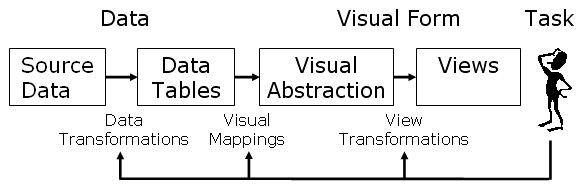}
  \caption{\label{fig:infovis_pipeline} The information visualization pipeline~\cite{Heer2006prefuse} depicting the data transformation and visual mapping process for constructing visualizations. }
  \vspace{-2em}
\end{figure}

Frameworks or pipelines for information visualization have been previously developed~\cite{Heer2006prefuse,Wijk05}. For example, the information visualization pipeline depicted in Figure 5 shows how data characteristics are extracted and assigned visual attributes or encodings, ultimately creating a visualization. The designs of visualizations adhering to this pipeline exhibit two primary components of the visual interface: the visualization showing the information, and a graphical user interface (GUI) consisting of graphical controls or widgets. The graphical controls in the GUI (e.g., sliders, knobs, etc.) allow users to directly manipulate the parameters they control. For example, ``direct manipulation''~\cite{Shneiderman1983Direct} user interfaces for information visualizations enable users to directly augment the values of data and visualization parameters to see the corresponding change in the visualization (e.g., using a slider to set the range of home prices and observing the filtering of results in a map showing homes for sale). This model is a successful user interaction framework for information visualizations.

Visual analytic systems have adopted this method for user interaction, but with the distinct difference of including analytic models or algorithms, as discussed earlier in this section. For example, in addition to filtering the data by selecting ranges for home prices, users could be given graphical controls over model parameters such as weighting the mixture of eigenvectors of a principal component analysis (PCA) dimension reduction (DR) model to produce two-dimensional views showing pairwise similarity of homes across all of the available dimensions. To users who lack expertise in such models, this may pose fundamental usability challenges. 

\begin{figure}
  \centering
  \includegraphics[width=0.99\columnwidth]{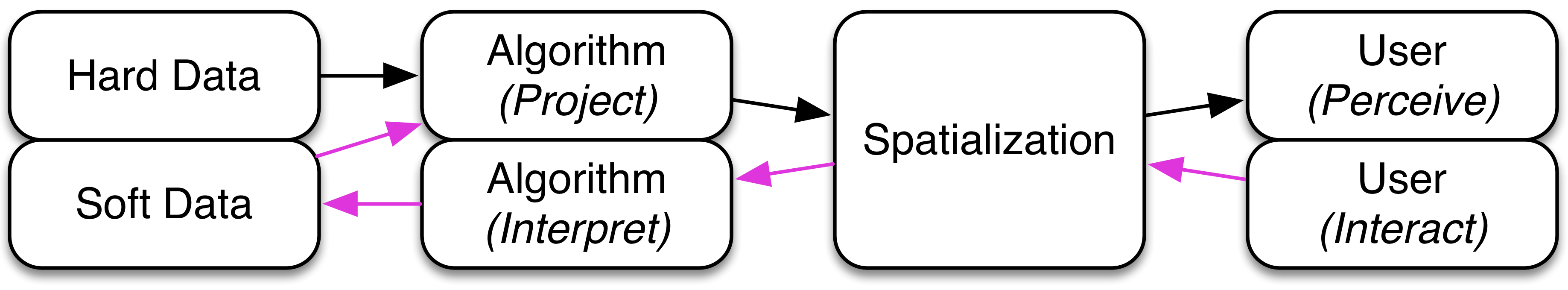}
  \caption{\label{fig:si_pipeline} The semantic interaction pipeline~\cite{Endert2012Semantic-chi} showing how the user interactions in a spatial visualization can be incorporated into the computation of a visual analytic system.}
  \vspace{-2em}
\end{figure}

In contrast, prior work has proposed frameworks to perform model steering via machine learning techniques applied to the user interactions performed during visual data analysis, called semantic interaction~\cite{Endert2012Semantic-chi}. Semantic interaction is an approach to user interaction for visual data exploration in which analytical reasoning of the user is inferred and in turn used to steer the underlying models implicitly (illustrated in Figure~\ref{fig:si_pipeline}). The goal of this approach to user interaction is to enable co-reasoning between the human and the analytic model (or models) used to create the visualization (coupling cognition and computation) without requiring the user to directly control the models.

The approach of semantic interaction is to overload the visual metaphor through which the insights are obtained (i.e., the visualization of information created by computational models) and the interaction metaphor through which hypotheses and assertions are communicated (i.e., interaction occurs within the visual metaphor). Semantic interaction enables users to directly manipulate data within visualizations, from which tacit knowledge of the user is captured, and the underlying analytic models are steered. The analytic models can be incrementally adapted based on the user's sensemaking process and domain expertise explicated via the user interactions with the system (as described in the models of Section 2.1).

The semantic interaction pipeline (shown in Figure~\ref{fig:si_pipeline}) takes an approach of directly binding model steering techniques to the interactive affordances created by the visualization. For example, a distance function used to determine the relative similarity between two data points (often visually depicted using distance in a spatial layout), can serve as the interactive affordance to allow users to explore that relationship. Therefore, the user interaction is directly in the visual metaphor, creating a bi-directional medium between the user and the analytic models~\cite{Leman2011Bi}. 

\subsection{Machine Learning Models and Frameworks}
There is not as much work in machine learning models and frameworks. Most of the proposals correspond to some form of \emph{de facto} industrial standards, such as the SEMMA (Sample, Explore, Modify, Model, and Assess) methodology advertised by SAS Institute Inc. Among those, a vendor neutral framework, CRISP-DM \cite{shearer2000crisp}, is somewhat comparable to knowledge discovery and visual analytics frameworks. There are six phases in the framework: business (or problem) understanding; data understanding (developed through exploration of the data and discussion with data owners); data preparation (including feature extraction, noise removal, and transformation); modeling; evaluation (testing the quality of the model, and particularly its generalization performance); deployment (embedding the model in practice). In some versions of this framework, there is an additional link from deployment back to business understanding - this represents the fact that the underlying data generator may change over time. The model needs continuous evaluation in deployment and when performance degrades, the process starts again. Perhaps more importantly, all the steps of the framework are embedded in a general loop comparable to the ones observed in other frameworks. This emphasize the feedback from the latter stage of the process (evaluation in numerous machine learning applications) to the early stages (e.g. data preparation in CRISP-DM).

As pointed out in e.g. \cite{amershi2014power}, the traditional implementation of the machine learning workflow leads to long development cycles where end users (who are also domain experts) are asked to give feedback on the modeling results. This feedback is used by machine learning experts to tune the whole processing chain, especially at the data preparation stage. Ideally, this feedback should take the form of specific and formal user inputs, for example positive and negative feedback on exemplars (such as ``those two objects should not belong to the same cluster'' or ``this object is misclassified'').

\begin{figure}
  \centering
  \includegraphics[width=0.65\columnwidth]{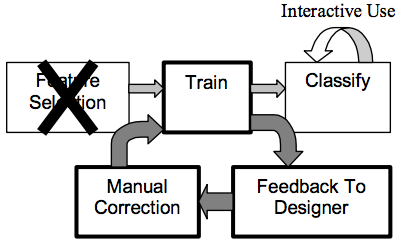}
  \caption{\label{fig:iml} A model for interactive machine~\cite{fails2003interactive} learning depicting user feedback for model training.}
  \vspace{-2em}
\end{figure}

User feedback in this formal, expressive form lends itself very well to steering and training machine learning models, for example via \emph{interactive machine learning} approaches \cite{PorterEtal2013IML}. Figure~\ref{fig:iml} shows an early model of interactive machine learning that emphasizes the feedback that users give to train classifiers~\cite{fails2003interactive}. Through multiple iterations of feedback, the classifier gets more training examples, and is thus able to more closely approximate the phenomena or concept being classified in the data.

To further establish an ML framework, we note the following. Machine learning tasks are traditionally divided into two broad categories, supervised tasks and unsupervised tasks. In supervised learning, the goal is to construct a model that maps an input to an output, using a set of examples of this mapping, the training set. The quality of the model is evaluated via a fixed loss criterion. Up till recently, it has generally been considered that human input is not needed in the model construction phase. On the contrary, it could lead to undetected overfitting. Indeed the expected quality of the model on future data (its so-called generalization ability) is generally estimated via an independent set of examples, the test set. Allowing the user (or a program) to tune the model using this set will generally reduce the generalization ability of the model and prevent any sound evaluation of this ability (unless yet another set of examples is available).

Supervision via examples can be seen as a direct form of user control over the training process. Allowing the user to modify the training set interactively provides an indirect way of integrating user inputs into the model construction phase. In addition, opportunities for user feedback and control are available before and after this modeling step (e.g., using the CRISP-DM phases). For instance, user feedback can be utilized at the feature selection, error preferences, and other steps. Leveraging those opportunities (including training set modification) has been the main focus of interactive machine learning approaches. For instance, tools such as the Crayons system from \cite{fails2003interactive} allow the user to add new training data by specifying in a visual way positive and negative examples. This specific type of user feedback in the form of labelling new examples is exactly the focus of the \emph{active learning} framework \cite{settlestr09} in machine learning. This learning paradigm is a variation over supervised learning in which ML algorithms are able to determine interesting inputs for which they do not know the desired outputs (in the training set), in such a way that given those outputs the predictive performances of the model would greatly improve. Interestingly active learning is not the paradigm used in e.g. \cite{fails2003interactive}. It seems indeed that in real world applications, active learning algorithms tend to ask too many questions and possibly to similar ones, as reported in e.g.,  \cite{ICML2011Guillory_261}. More generally, the need for specific and formal user inputs can create usability issues with regards to people and their tasks, as pointed out in e.g.,  \cite{amershi2014power,Endert2014human}. That is, the actions taken by the user to train the systems are often not the actions native to the exploratory data analysis described in the previously mentioned frameworks. This is starting to become more commonly used in the ML community, as exemplified by \cite{BalcanHanneke2012}. In this paper the authors consider additional questions a system can ask a user, beyond just labelling. They focus in particular on \emph{class conditional queries} -- the system shows the user unlabeled examples and asks him or her to select one that belongs to a given class (if one exists).

In unsupervised learning, the data has no input/output structure and the general goal is to summarize the data in some way. For instance, as discussed further below, dimension reduction techniques build low dimensional approximations of the data from their high dimensional initial representation; clustering groups data into similar objects; etc. Unsupervised learning is generally considered ill posed in the ML field in the following sense: most of the tasks of unsupervised learning (clustering, dimensionality reduction, etc.) have only an informal description to which numerous formal models can be related. Those models are very difficult to compare on a theoretical point of view as well as on a practical one. In unsupervised learning, the need for user input, steering and control is therefore broadly accepted and techniques to include user feedback into e.g., clustering have been studied for some time. Variations over unsupervised methods that take explicitly into account some form of additional information are generally called semi-supervised methods. The supervision is frequently provided by external data in an automated way, but those methods can lead to principled ways of integrating user feedback.

It should be noted however that most of the methodological development in machine learning that can be used to integrate user feedback, from active learning to triplet based constraints \cite{vanDerMaatenWeinberger2012triplet}, are seldom evaluated in the context of visualization systems. In general, the feedback process is either simulated or obtained via off line and slow process (e.g. Amazon's Mechanical Turk for triplet in \cite{WilberEtAl2015}). Thus while specific frameworks that enable user feedback have been defined by the ML community, the practical relevance of the recent ones in the context of interactive visualization remains untested.

\subsection{Comparison to another classification framework}
A recent paper by Sacha et al.~\cite{sacha2016visual} overlaps with this STAR Report. 
It focuses on the specific area of dimensionality reduction and how these techniques integrate with interactive visualization in visual analytics systems. 
The paper builds around a systematic analysis of visualization literature, which reveals seven common interaction scenarios. 
The evaluation leads to the identification of future research opportunities.

The current paper provides a significantly broader survey of machine learning methods coupled with interaction, while Sacha et al.~\cite{sacha2016visual} probe deeper in one important area. 
In addition to dimension reduction, the current paper deals with ML methods for clustering, classification, and regression. 
There is some overlap in the literature covered in the two papers. 
However, the literature reviewed in the current paper cites ML methods that are already coupled with interactive visualization systems plus those that are not yet (but it would be beneficial if they were); Sacha et al. deal mostly with ML methods that are already coupled with interactive visualizations.

The two papers complement each other with Sacha's deeper analysis in DR strengthening the wider analysis in the current paper, and vice versa. 
The human-in-the-loop process model in~\cite{sacha2016visual} has similarities with the use of the human-machine interaction loop in the current paper; they also share a common origin. 
The classifications used in Sacha et al's structured analysis are different than those in the current paper's taxonomy, although one could be mapped into the other, with modifications. 
However, there are also multiple similarities; in particular, classification according to ``modify parameters and computation domain'' and ``define analytical expectations'' in Sections~\ref{sec:modify_parameters} and~\ref{sec:define_analytical_expectations} of the current paper map to various interaction scenarios in Sacha et al.~\cite{sacha2016visual}. 
For example, the first classification maps to data manipulation, DR parameter tuning, and DR type selection scenarios in Sacha et al's model. 
The second classification, in permitting the user to tell the system (based on results it gives) expectations that are consistent with domain knowledge, maps to feature selection and emphasis and defining constraints scenarios. 
The current paper then goes beyond DR, including for each classification a discussion of clustering, classification, and regression methods. 
This broadens and strengthens the discussion from Sacha et al.~\cite{sacha2016visual}.

\section{Categorization of Machine Learning Techniques Currently used in Visual Analytics}
\label{sec:category}

The visual analytic community has developed systems that leverage specific machine learning techniques. In this section, we give an overview of the existing ways that machine learning has been integrated into VA applications from two transversal perspectives: the \emph{types of ML algorithms} and the so-called \emph{interaction intent}. We pay special attention to the ``interaction intent'' as described below, because this focuses on human-in-the-loop aspects that are central to VA systems. There are also other papers where the main role of visualization is on communicating the results of computations to improve comprehension~\cite{turkay2014computationally} that are not directly covered in this section. Some of the most significant of these papers, referring to VA systems, are described in Section~\ref{sec:domains}.

Along the first perspective, we consider the different \textit{types of ML algorithms} that have been considered within visual analytics literature. Although one might think of several other possible ways to categorize the algorithms~\cite{alpaydin2014introduction, friedman2001elements}, here we adopt a high-level task-oriented taxonomy and categorize the algorithms under the following headings: \textit{dimension reduction}, \textit{clustering}, \textit{classification}, \textit{regression/correlation analysis}. We observe that ML algorithms to tackle these tasks are frequently adopted in visual analytics applications since these analytical tasks often require the joint capabilities of computation and user expertise. To briefly summarize: i) \textit{dimension reduction} methods help analysts to distill the information in high-dimensional data so that conventional visualization methods can be employed and important features are identified ii) \textit{clustering} methods help to identify groups of similar instances which can be done both in a supervised or unsupervised manner iii) \textit{classification} methods are often supervised and help to build models to associate labels to data instances, and finally iv) \textit{regression/correlation} analysis methods help to investigate relations between features in the data and to understand/generate causal links to explain phenomena.  
  
Along the second perspective, we focus on the user side of the process. We name this aspect as \textit{interaction intent} and categorize the actions taken by users within visual analysis in terms of the methods through which the analyst tries to improve the ML result. 

This perspective of our taxonomy resonates with the ``\textit{user intent}'' categories suggested by Yi et al.\cite{yi2007toward} for low-level interactions within InfoVis applications. Our focus, however, is targeted on higher-level analytical intents within the narrower scope of visual analytics applications that involve ML methods. With this motivation in mind, we suggest two broad categories for ``\textit{intents}'': \textit{modify parameters and computation domain} and \textit{define analytical expectations}. Table 1 shows the organization of literature along the dimensions of algorithm type vs the two categories of user intent. Here we survey the existing literature within the scope of this characterization.

\subsection{Review Methodology}
The literature summarized and categorized in this section are taken from impactful ML and visualization conferences and journals. They were chosen and categorized based on discussions the authors had at the Dagstuhl Seminar titled, ``Bridging Machine Learning with Information Visualization (15101)'' \cite{keim_et_al:DR:2015:5266}, and later refined through a more extensive literature review.

Within this report, we review existing literature on the integration of machine learning and visualisation from three different perspectives -- models and frameworks, techniques, and application areas.  
When identifying the relevant works in these domains, we follow a structured methodology and identified the different scopes of investigation for these three different perspectives. 
One important note to make is, due to our focus on the integration of the two fields, we scanned resources from both the visualisation and machine learning domain.

Within the domain of visualisation, we initiated our survey starting with publications from the following resources: 

\noindent\emph{\textbf{Journals:}}  IEEE Transactions on Visualization and Computer Graphics, Computer Graphics Forum, IEEE Computer Graphics and Applications, Information Visualization 

\noindent\emph{\textbf{Conferences:}} IEEE Visual Analytics Science and Technology (partially published as a special issue of IEEE TVCG), IEEE Symposium on Information Visualization (InfoVis) (published as a special issue of IEEE TVCG since 2006),  IEEE Pacific Visualization Symposium (PacificVis), EuroVis workshop on Visual Analytics (EuroVA)

Within the domain of machine learning, we initiated our survey starting with publications from the following resources: 

\noindent\emph{\textbf{Journals:}}  Journal of Machine Learning Research, Neurocomputing, IEEE Transactions on Knowledge and Data Engineering 

\noindent\emph{\textbf{Conferences:}} International Conference on Machine Learning (ICML), ACM SIGKDD International Conference on Knowledge Discovery and Data Mining, European Symposium on Artificial Neural Networks, Computational Intelligence and Machine Learning (ESANN)

We then scanned the relevant papers identified in the above resources and performed a backward and forward literature investigation using Google Scholar. 
In producing the taxonomy of works within Section~\ref{sec:category}, we labelled the publication both in terms of the analytical task and the integration strategy incorporated.

\begin{table*}[t!]
\centering
\begin{tabular}{ p{1.5cm} p{6cm}p{6cm}}
	
	 & Modify Parameters \& \par Computation Domain & Define \par Analytical Expectations \\
	\toprule
   Dimension \par Reduction & ~\cite{johansson2009interactive},~\cite{Fernstad2011},~\cite{fuchs2009},~\cite{stahnke2016probing}, \par ~\cite{Williams2004},~\cite{nam2013tripadvisor},~\cite{turkay2011brushing},~\cite{turkay2012representative}, \par ~\cite{janicke2008brushing},~\cite{amir2013visne},~\cite{Jeong09}    & ~\cite{endert2011observation}, ~\cite{endert2013beyond},~\cite{brown2012dis}, \par ~\cite{hu2013semantics},~\cite{garg2008model},~\cite{conf/aistats/IwataHG13}, ~\cite{Kaski11}, \par ~\cite{Perez15},~\cite{Kim2016Interaxis},~\cite{kwon2016axisketcher}  \\
   \midrule
	Clustering  & ~\cite{kandogan2012just},~\cite{rinzivillo2008},~\cite{schreck2008},~\cite{rasmussen2004}, \par ~\cite{seo2002},~\cite{lex_stratomex:_2012},~\cite{lex_comparative_2010},~\cite{turkay2014characterizing}, \par ~\cite{turkay2011integrating},~\cite{Ahmed2012},~\cite{rinzivillo2008},~\cite{hadlak2013supporting}, \par ~\cite{turkay11temporalCluster},~\cite{parulek2013visual},~\cite{Henry2013MultiFacet},~\cite{Wise1999Visualizing}, \par ~\cite{younesy2013interactive},~\cite{streit2014furby}   &~\cite{hossain2012scatter},~\cite{Choo13},~\cite{basu2008constrained},~\cite{cohn2008semi},\par ~\cite{BasuEtAl2004Aactive},~\cite{AwasthiEtal2014},~\cite{KumarEtAl2004}, ~\cite{KumarKummanmuru2008} \\
	\midrule
	Classification &~\cite{DBLP:conf/aaai/PoulinESLGWFPMA06},~\cite{may2008towards},~\cite{may2011guiding},~\cite{van2011baobabview}, \par\cite{choo2010ivisclassifier},  ~\cite{krause2014infuse},~\cite{andrienko2010space},~\cite{andrienko2009interactive},\par \cite{klemmivapp15} & ~\cite{settlestr09},~\cite{DBLP:journals/jmlr/StrumbeljK10},~\cite{behrisch2014feedback},~\cite{paiva2015approach}\\
	\midrule
	Regression & ~\cite{Piringer2010HyperMoval},~\cite{muhlbacher2013partition},~\cite{malik_correlative_2012},~\cite{turkay2012representative}, \par ~\cite{klemm20163d} & ~\cite{matkovic2008interactive},~\cite{matkovic2014visual}~\cite{Lu2014Integrating}~\cite{yeon2016predictive}\\

	\bottomrule \\
	\end{tabular}
	\caption{In Section~\ref{sec:category}, we review the existing literature in visual analytics following a 2D categorization that organizes the literature along two perspectives: Algorithm Type (rows) and Interaction Intent (columns). }
	\label{tab:taxonomy}
	\vspace{-2em}
\end{table*}

\subsection{Modify parameters and computation domain} 
\label{sec:modify_parameters}
Here we list techniques where interaction has been instrumental in modifying the parameters of an algorithm, defining the measures used in the computations, or even changing the algorithm used. Another common form of interaction here is to enable users to modify the computational domain to which the algorithm is applied. Such operations are often facilitated through interactive visual representations of data points and data variables where analysts can select subsets of data and run the algorithms on these selections within the visual analysis cycle to observe the changes in the results and to refine the models iteratively. The types of techniques described in this section can be considered as following a ``direct manipulation''~\cite{Shneiderman1983Direct} approach where the analysts explicitly interact with the algorithm before or during the computation and observe how results change through visualization.

\paragraph*{Dimension Reduction} 
One class of algorithms that is widely incorporated in such explicit modification strategy is dimension reduction. Since high-dimensional spaces are often cognitively challenging to comprehend, combinations of visualization and dimension reduction methods have demonstrated several benefits.  
Johansson and Johansson~\cite{johansson2009interactive} enable the user to interactively reduce the dimensionality of a data set with the help of quality metrics. The visually guided variable ordering and filtering reduces the complexity of the data and provides the user a comprehensive control over the whole process. The authors later use this methodology in the analysis of high-dimensional data sets involving microbial populations~\cite{Fernstad2011}. An earlier work that merges visualization and machine learning approaches is by Fuchs et al.~\cite{fuchs2009}. The authors utilize  machine learning techniques within the visual analysis process to interactively narrow down the search space and assist the user in identifying plausible hypotheses. In a recent paper, Stahnke et al.~\cite{stahnke2016probing} devised a probing technique using interactive methods through which analysts can modify the parameters of a multi-dimensional scaling projection. The visualization plays a key role here to display the different dimension contributions to the projections and to communicate the underlying relations that make up the clusters displayed on top of the projection results.

In MDSteer~\cite{Williams2004}, an embedding is guided by user interaction leading to an adapted multidimensional scaling of multivariate data sets. Such a mechanism enables the analyst to steer the computational resources accordingly to areas where more precision is needed. This technique is an early and good example of how a deep involvement of the user within the computational process has the potential to lead to more precise results. Nam and Mueller~\cite{nam2013tripadvisor} provide the user with an interface where a high-dimensional projection method can be steered according to user input. They provide ``key'' computational results to guide the user to other relevant results through visual guidance and interaction. Turkay et al. introduce the dual-analysis approach~\cite{turkay2011brushing} to support analysis processes where computational methods such as dimension reduction~\cite{turkay2012representative} are used. The authors incorporate several statistical measures to inform analysts on the relevance and importance of variables. They provide several perspectives on the characteristics of the dimensions that can be interactively recomputed so that analysts are able to make multi-criteria decisions whilst using computational methods. J\"{a}nicke et al.~\cite{janicke2008brushing} utilize a two-dimensional projection method where the analysis is performed on a projected 2D space called the attribute cloud. The resulting point cloud is then used as the medium for interaction where the user is able to brush and link the selections to other views of the data. In these last group of examples, the capability to run the algorithms on user-defined subsets of the data through visually represented rich information is the key mechanism to facilitate better-informed, more reliable data analysis processes.

\paragraph*{Clustering} Clustering is one of the most popular algorithms that have been integrated within visual analytics applications. Since visual representations are highly critical in interpreting and comprehending the characteristics of clusters produced by the algorithms, direct modification of clustering algorithms are often facilitated through interactive interfaces that display new results ``on-demand''. gCluto~\cite{rasmussen2004} is an interactive clustering and visualization system where the authors incorporate a wide range of clustering algorithms. This is an early example where multiple clustering algorithms can be run on-the-fly with varying parameters and results can be visually inspected.
In \emph{Hierarchical Clustering Explorer}~\cite{seo2002}, Seo and Shneiderman describe the use of an interactive dendogram coupled with a colored heatmap to represent clustering information within a coordinated multiple view system.

\begin{figure}[t]
  \centering  
  \includegraphics[width=0.8\columnwidth]{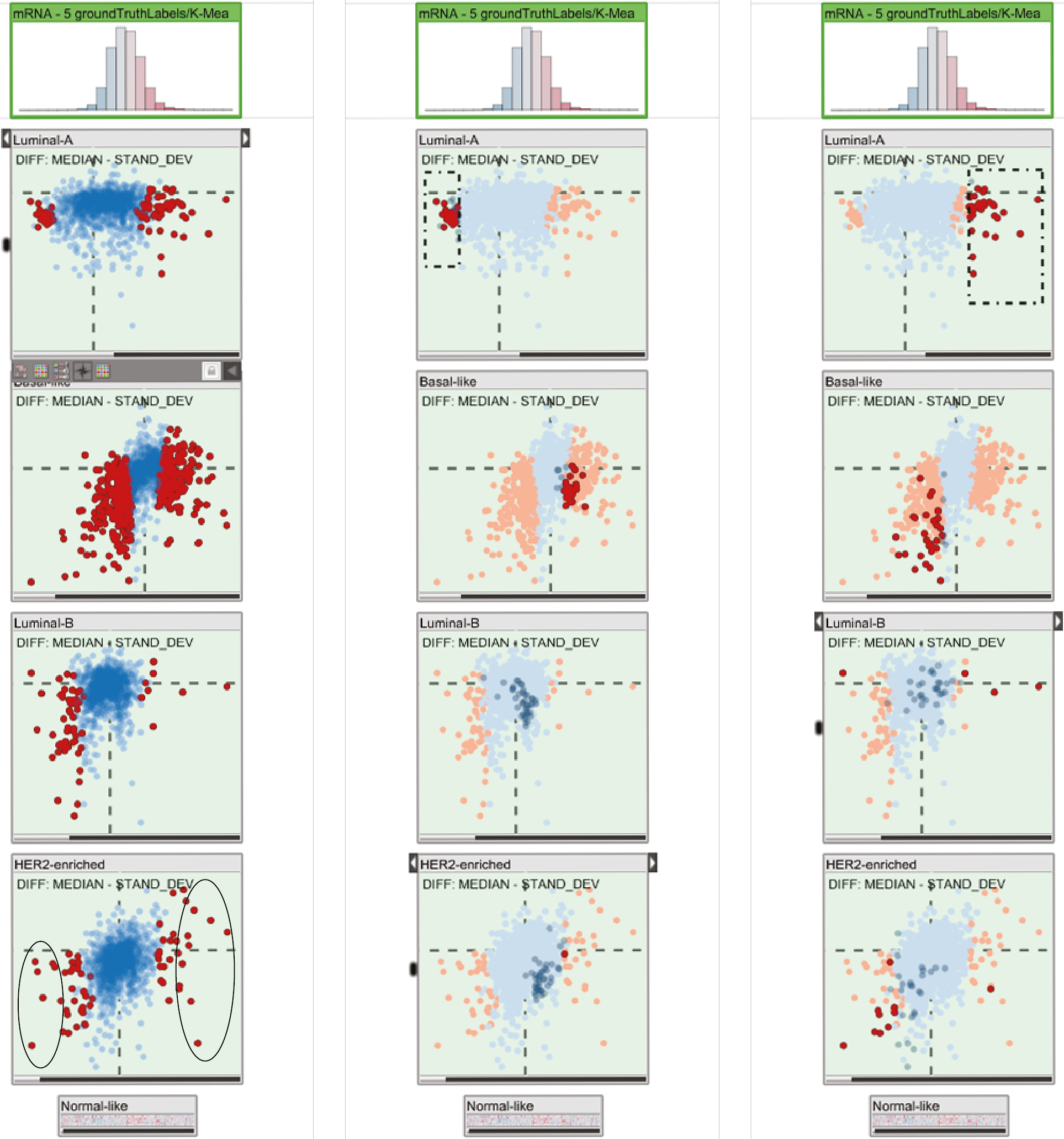}
  \caption{\label{fig:turkayClustering} Visualization of clustering results, together with associated on-the-fly computations to identify discriminating features of groups, are used here to aid analysts in interpreting the clusters and refining them further~\cite{turkay2014characterizing}. }
  \vspace{-2em}
\end{figure}

Other examples include work accomplished using the Caleydo software for pathway analysis and associated experimental data by Lex et al.~\cite{lex_stratomex:_2012, lex_comparative_2010}. In their techniques, the authors enable analysts to investigate multiple runs of clustering algorithms and utilize linked, integrated visualizations to support the interpretation and validation of clusters. Along the same lines, Turkay et al. present an interactive system that addresses both the generation and evaluation stages within the clustering process and provides interactive control to users to refine grouping criteria through investigations of measures of clustering quality~\cite{turkay2011integrating}. In a follow-up work~\cite{turkay2014characterizing}, within the domain of clustering high-dimensional data sets, integrated statistical computations are shown to be useful to characterize the complex groupings that analysts encounter in such data sets. Figure~\ref{fig:turkayClustering} demonstrates how the authors incorporated statistical analysis results to indicate important features for data groups. In this work, the most discriminative features (indicated with red dots as opposed to blue ones that are less important) for the clustering result of a high-dimensional data set are represented as integrated linked views. The user is able to select these features in one clustering result (e.g., within the clustering result in the right-most column in Figure~\ref{fig:turkayClustering}) and observe whether the same features are represented in others, e.g., in the left-most column.

Schreck et al.~\cite{schreck2008} propose a framework to interactively monitor and control Kohonen maps to cluster trajectory data. The authors state the importance of integrating the expert within the clustering process for achieving good results.  Kandogan~\cite{kandogan2012just} discusses how clusters can be found and annotated through an image-based technique. His technique involves the use of ``just-in-time'' clustering and annotation, and the principal role for visualisation and interaction is to aid the interpretation of the structures observed, and provide a deeper insight into why and how particular structures are formed.

An important role for visualization is to get the user engaged in \textit{progressive} and \textit{iterative} generation of clusters~\cite{rinzivillo2008}. In such approaches, the user is presented with content that is built step-by-step and gains additional insight in each iteration to decide whether to continue, alter, or terminate the current calculations. Such levels of interactivity, of course, require the solutions to be responsive and capable of returning results within acceptable delays. Ahmed and Weaver~\cite{Ahmed2012} address this problem through forward-caching expected interaction possibilities and providing users with clustering results without breaking the responsive analytical flow.

Visual analytics applications that involve clustering algorithms within the analysis of complex dynamic networks have also been developed~\cite{hadlak2013supporting}. The use of visualisation is in particular critical with such dynamic relational data sets due to the limitations in interpreting the algorithmic results; well-designed combinations of visual summaries can assist analysts in this respect. In the domain of molecular dynamics simulation, there are some examples of tight integrations of interactive visualizations, clustering algorithms, and statistics to support the validity of the resulting structures~\cite{turkay11temporalCluster},~\cite{parulek2013visual}.

\paragraph*{Classification} 

Being a relevant and widely utilized technique, classification algorithms have also found their place within visual analytics applications. Common roles for interactive visualization are filtering the feature space, iteratively observing and fixing problems, and when the classification tasks involve multiple mediums such as space, time and abstract features, providing multiple perspectives to the algorithmic results.

A conceptual framework on how classification tasks can be supported by interactive visualizations is presented by May and Kohlhammer~\cite{may2008towards}. Their approach improved the classification of data using decision trees in an interactive manner. They proposed the use of a technique called KVMaps to inform users on classification quality thus enabling the iterative refinement of the results. The authors later proposed a technique called SmartStripes~\cite{may2011guiding} where they investigated the relations between different subsets of features and entities. Interactive visual representations have been used to help create and understand the underlying structures within decision trees~\cite{van2011baobabview}. The authors not only presented the overall structure of decision trees, but also provided intuitive visual representations of attribute importance within the different levels of the tree. Such interactive visualizations are critical in unraveling the computed information hidden within the layers and can be quite instrumental in increasing the trust in such computational models. Similar insights can be gained on other models (additive ones, e.g. naive Bayes, in \cite{DBLP:conf/aaai/PoulinESLGWFPMA06} and more general ones in \cite{DBLP:journals/jmlr/StrumbeljK10}) by \emph{explaining} individual classification. In these papers, the authors display the contribution of features to the classification made by the model and enable what-if scenarios, such ``how would the classification change if this particular feature was set to another value?'' 

In iVisClassifier by Choo et al.~\cite{choo2010ivisclassifier}, the authors improve classification performance through interactive visualizations. Their technique supports a user-driven classification process by reducing the search space, e.g., through recomputing Latent Dirichlet Allocation (LDA)~\cite{blei2003latent} with a user-selected subset of data defined through filtering in additional coordinated views.  Klemm et al.~\cite{klemmivapp15} investigate the use of interactive visualisation to compare multiple decision trees in investigating relations within non-image and image based features for a medical application. They visualise the quality aspects of classifiers to infer observations on the predictive power of the features. 

\begin{figure}[t]
  \centering
  \includegraphics[width=0.99\columnwidth]{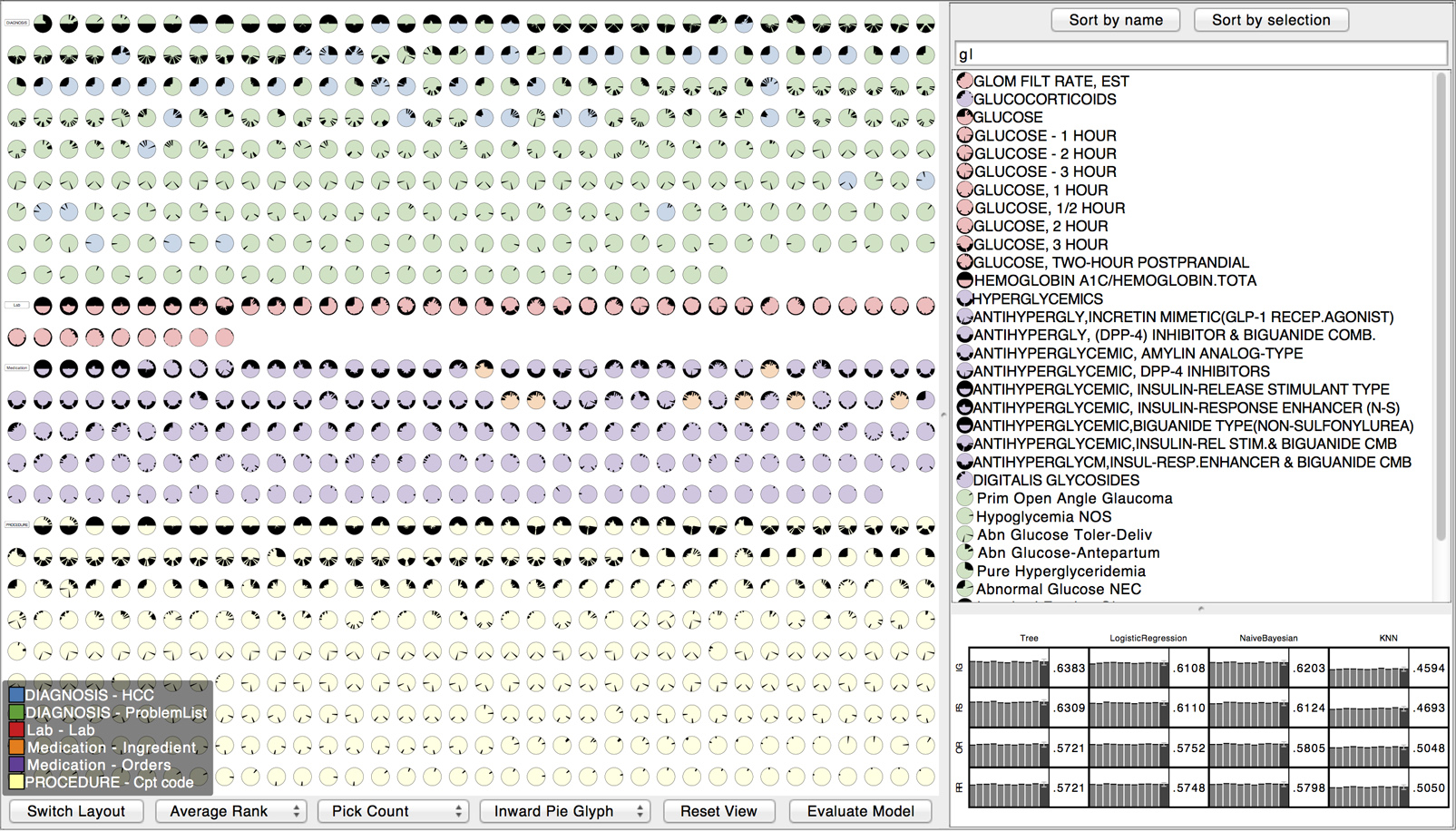}
  \caption{\label{fig:infuse} Visual summaries to indicate the relevance of features over cross-validation runs support analysts in making informed decisions whilst selecting features for a classification model~\cite{krause2014infuse}.}
  \vspace{-2em}
\end{figure}

Krause et al.~\cite{krause2014infuse} address the important process of feature selection within model building for classification purposes. Through visual representations of cross-validation runs for feature ranking with various algorithms, their method supports the decisions made while including or excluding particular features from a classification model (see Figure~\ref{fig:infuse}). Their approach enables users to be part of the predictive model building process and, as also demonstrated by the authors, leads to better performing/easier to interpret models. Their methodology is based on producing glyphs for the features of a data set to represent how important each one is within a number of classification models. In addition, the glyphs are also used as elements for visual selections and enable analysts to interactively apply modelling on subsets of features.

Classification of spatio-temporal patterns is one of the complex tasks that requires the involvement of user input and efficient algorithms due to the complex nature of structures found in such data sets. Andrienko et al.~\cite{andrienko2010space} investigate how self organizing maps (SOMs) are integrated into the visual analysis process. They integrate a SOM matrix where the user can interactively modify the parameters and observe the changes in the results in various visual representations, e.g., where space is represented in time, and the time is represented in space. Again involving spatio-temporal data, an interactive process where a clustering algorithm assists users to pick relevant subsets in building classifiers has shown to be effective in categorizing large collections of trajectories~\cite{andrienko2009interactive}.   

\paragraph*{Regression} 
Identifying the multivariate relations within data variables, in particular when their numbers are high, is one of the critical tasks in most data analysis routines. In order to evaluate to what degree observed relations can be attributed to underlying phenomena and to build causal interpretations, visual analytics approaches have shown good potential. 
Visualization has shown to be effective in validating predictive models through interactive means~\cite{Piringer2010HyperMoval}. The authors visually relate several n-dimensional functions to known models through integrated visualizations within a model building process. They observed that such a visualization-powered approach not only speeds up model building but also increases the trust and confidence in the results.  M\"{u}hlbacher and Piringer~\cite{muhlbacher2013partition} discuss how the process of building regression models can benefit from integrating domain knowledge. Berger et al.~\cite{berger2011} introduce an interactive approach that enables the investigation of the parameter space with respect to multiple target values.
Malik et al.~\cite{malik_correlative_2012} describe a framework for interactive auto-correlation. This is an example where the correlation analysis is tightly coupled with the interactive elements in the visualization solution. Correlation analysis has been integrated as an internal mechanism to investigate how well lower-dimensional projections relate to the data that they represent~\cite{turkay2012representative}. The use of relational representations here supports analysts to evaluate how local projection models behave in preserving the correlative structures in the data. In a recent paper, Klemm et al.~\cite{klemm20163d} demonstrates the use of visualisation to show all combinations of several independent features with a specific target feature. The authors demonstrate how the use of template regression models, interactively modifiable formulas and according visual representations help experts to derive plausible statistical explanations for different target diseases in epidemiological studies.

\subsection{Define analytical expectations} 
\label{sec:define_analytical_expectations}
Unlike the papers in the previous category where the user explicitly modifies the parameters and the settings of an algorithm, the works we review under this section follow a different strategy and involve users in communicating \textit{expected result}s to the computational method. In these types of interactive methods, the user often observes the output of an algorithm and tell the machine which aspect of the output is inconsistent with the existing knowledge, i.e., correcting the algorithm. Furthermore, analysts can also communicate examples of relevant, domain-knowledge informed relations to be preserved in the final result. Since this is a relatively recent approach to facilitate the interaction between the user and the algorithms, the number of works in this category is not as high as the previous section. In the following, we review such works again under a categorization of different ML algorithm types involved. Notice that integrating user knowledge in this way in unsupervised learning contexts falls into the general semi-supervised framework, which is a principled way in ML for making unsupervised problems less ill-posed.

\paragraph*{Dimension Reduction}
Dimension reduction algorithms are suitable candidates for such approaches due to the often ``unsupervised'' nature of the algorithms and the possibility that errors and losses within the reduction phase are high, in particular with datasets with high numbers of dimensions. As one of the early works along these lines, Endert et al.~\cite{endert2011observation} introduce observation level interactions to assist computational analysis tools to deliver more interpretable/reliable results. The authors describe such operations as enabling the \textit{direct manipulation} for visual analytics~\cite{endert2013beyond}. In this line of work, the underlying idea is to provide mechanisms to users to reflect their knowledge about the data through interactions that directly modify computational results. One typical interaction is through \textit{moving} observations in a projection such that the modified version is more similar to the \textit{expectation} of the analyst~\cite{endert2011observation,brown2012dis}. This line of research has been expanded to focus on the interpretability of linear~\cite{Kim2016Interaxis} and non-linear DR models~\cite{kwon2016axisketcher}. Hu et al.~\cite{hu2013semantics} complemented such visualization level interaction methods with further interaction mechanisms. The authors aim to understand users' interaction intent better and give them mechanisms to also highlight preferences on \textit{unmoved} points. 

In their Model-Driven Visual Analytics system, Garg et al.~\cite{garg2008model} suggest the use of a "pattern painting" mechanism that enables analysts to paint interesting structures in the visualization which are then turned into logical rules that can be fed into a projection algorithm to build an effective model. 

An interesting supervised point of view has been proposed in \cite{conf/aistats/IwataHG13} on the dimension reduction steering. The main idea is to introduce an information theoretic criterion that evaluates the uncertainty in the representation, considering that the original high dimensional points are noisy. Given this criterion, the authors apply an active learning approach to select points that are maximally informative: if the user can move one of those points to its desired position, the uncertainty of the representation will be maximally reduced (compared to the reduction expected with other points). The experimental evaluation shows that the optimal points tend to be more uniformly distributed over the projected data set than with other selection methods, possibly reducing some of the drawbacks of active learning summarized in e.g. \cite{amershi2014power}.

\begin{figure}[t]
  \centering
  \includegraphics[width=0.90\columnwidth]{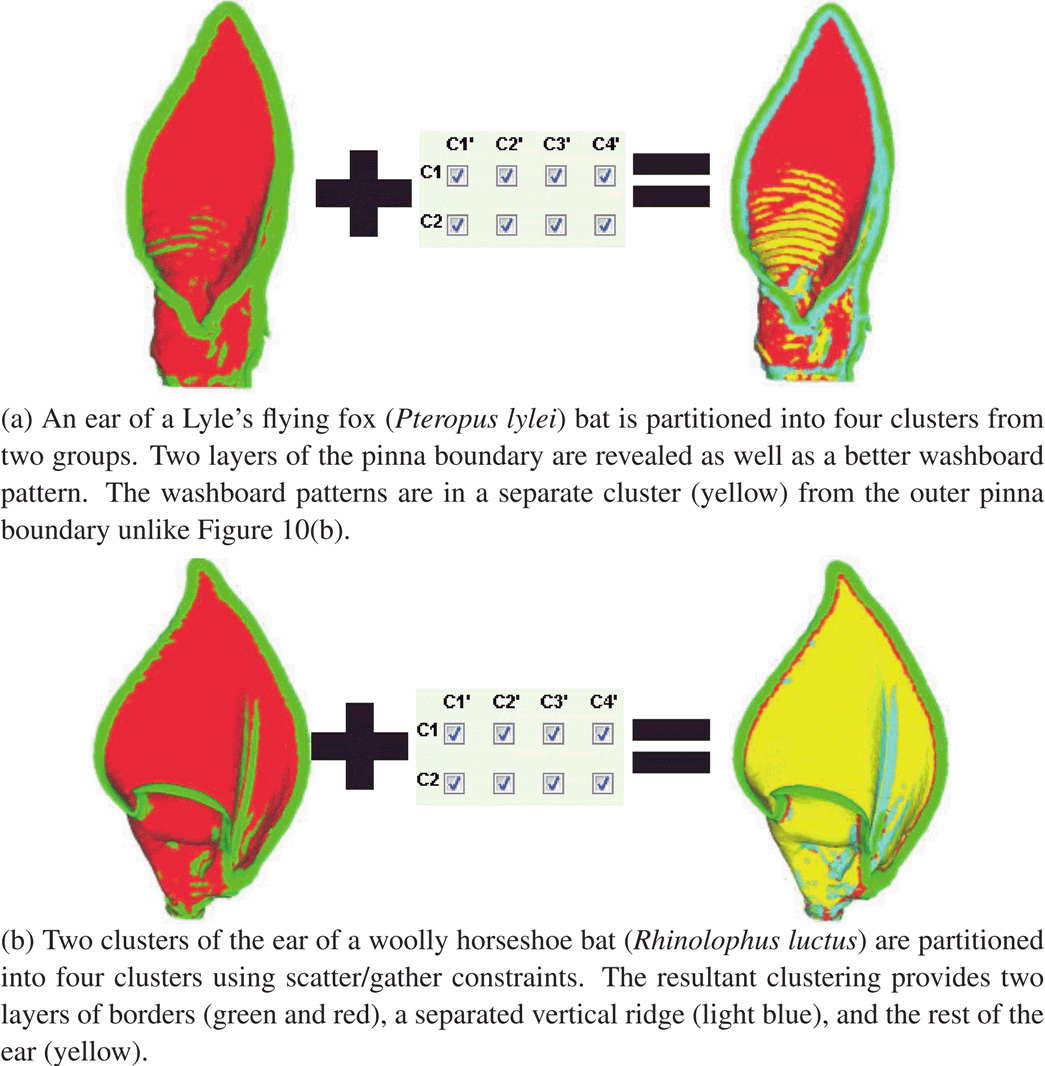}
  \caption{\label{fig:scatterGather} Scatter\/Gather~\cite{hossain2012scatter} is a technique to interactively gather feedback from analysts in response to algorithmic output and refine user-generated constraints to improve the clustering.}
  \vspace{-2em}
\end{figure}

\paragraph*{Clustering} 
There are a number of works where user knowledge is incorporated to feed a clustering algorithm with expected results. Hossain et al. makes use of a scatter\/gather technique to iteratively break up or merge clusters to generate groupings that meet analysts' expectations~\cite{hossain2012scatter}. (See Figure~\ref{fig:scatterGather}.) In their technique, the expert iteratively introduces constraints on a number of required relations and the algorithms take these constraints into consideration to generate more effective groupings. The users state whether clusters in the current segmentation should be broken up further or brought back together. Upon inspection of a clustering result, the user interactively constructs a scatter gather constraint matrix which represents a preferred clustering setting from her perspective. The algorithm then considers this input along with the clustering result to come up with an ``optimized'' result. In a number of papers, the user has been involved even further to modify clustering results. In order to support a topic modeling task through clustering, Choo et al.~\cite{Choo13} enable users to interactively work on topic clusters through operations such as splitting, merging and also refining clusters by pointing to example instances or keywords. 

More generally, clustering is one of the first tasks of machine learning to include ways to take into account expert knowledge, originally in the form of contiguity constraints (see \cite{murtagh1985survey} for an early survey): the expert specifies a prior neighborhood structure on data points (for instance related to geographical proximity) and the clusters are supposed to respect this structure (according to some notion of agreement). While the original methodology falls typically into the offline slow steering category, it has been extended to more general and possibly online steering based on two main paradigms for constraints clustering \cite{basu2008constrained}: the pairwise paradigm (with \emph{must-link}/\emph{cannot-link} constraints) and the triplet paradigm (with constraints of the form \emph{$x$ must be closer to $y$ than to $z$}). 

An early example of the pairwise paradigm is provided by \cite{cohn2008semi}. The authors describe a document clustering method that takes into account feedback of the form: this document should not belong to this cluster, this document should be in this cluster, those two documents should be (or should not be) in the same cluster (this mixes pointwise constraints, with pairwise ones). Active learning has been integrated into this paradigm in \cite{BasuEtAl2004Aactive}. A variation over the pairwise approach which consists in issuing merge and/or split requests at the cluster level has been proposed and studied in \cite{AwasthiEtal2014}.

Constraints based on triplet are more recent and were proposed in the context of clustering by \cite{KumarEtAl2004,KumarKummanmuru2008}. The main advantage of specifying triplet based constraints over pairwise ones is that they allow relative qualitative feedback rather than binary ones. They are also known to be more stable than pairwise comparisons \cite{gibbons1990rank}.

\paragraph*{Classification} 
Classification tasks are suitable for methods where users communicate known/expected/wrong classification results back to the algorithm. The ideas employed under this section show parallels to the Active Learning methodologies develop in the ML literature~\cite{settlestr09} where the algorithms have capabilities to query the user for intermediate guidance during the learning process.
In their visual classification methodology, Paiva et al.~\cite{paiva2015approach} demonstrates that effective classification models can be built when users' interactive input, for instance, to select wrongly labeled instances, can be employed to update the classification model. Along the similar lines, Behrisch et al.~\cite{behrisch2014feedback} demonstrate how users' feedback on the relevance of features in classification tasks can be incorporated into decision making processes. They model their process in an iterative dialogue between the user and the algorithm and name these stages as \textit{relevance feedback} and \textit{model learning}. This work serves as a good example of how user feedback might lead to better performing, fit-for-purpose classification models.

\paragraph*{Regression}

Although examples in this category are limited in numbers, defining the ``expected'' has shown great potential to support interactive visual steering within the context of ensemble simulation analysis~\cite{matkovic2008interactive,matkovic2014visual}. In their steerable computational simulation approach, Matkovic et al.~\cite{matkovic2008interactive} demonstrate how a domain expert (an engineer) can interactively define and refine desired simulation outputs while designing an injection system. Their three-level steering process enables the expert to define desired output values through selections in multiple views of simulation outputs. The expert then moves on to visually explore the control variables of the simulation and assess whether they are feasible and refine/re-run the simulation models accordingly. The authors went on to incorporate a regression model within this process to further optimise the simulation results based on users' interactive inputs~\cite{matkovic2014visual}. With this addition to the workflow, the experts again indicate desired output characteristics visually and a regression model followed by an optimization supports the process to quickly converge to effective simulation parameters. The critical role that the users play in these examples is to express their expert knowledge to identify and communicate suitable solutions to the algorithmic processes which in turn try and optimize for those.

\section{Application Domains}
\label{sec:domains}
The integration of ML techniques into VA systems has been exemplified in different domains, described below. Each of these domains present unique and important challenges, thus different combinations of interactive visualizations and ML techniques are used. Some of these techniques are related to, but go beyond the classifications in Section~\ref{sec:category}. For instance, dimension reduction, clustering, etc. since they must be closely embedded in the VA system and can be attached to higher level meanings. However, most are relevant to the Define Analytical Expectations category in Table~\ref{tab:taxonomy}. The examples given in this section generally make use of one or more technique categories in Section~\ref{sec:category}, depending on the particular domain for which the applications are designed for. 

\subsection{Text Analytics and Topic Modeling}
Text corpora are frequently analyzed using visual analytic systems. Text is a data format that lends itself nicely to specific computational processes, as well as human reasoning. Various text analytics methods have seen a lot of use in visual analytics systems over the past 6-7 years. A main reason is that these methods have proved useful in organizing large, unstructured text collections around meaningful topics or concepts. The text collections considered have been diverse including research publications, Wikipedia entries, streaming social media such as Twitter, Facebook entries, patents, technical reports, and other types. 

\begin{figure*}
  \centering
  \includegraphics[width=0.8\linewidth]{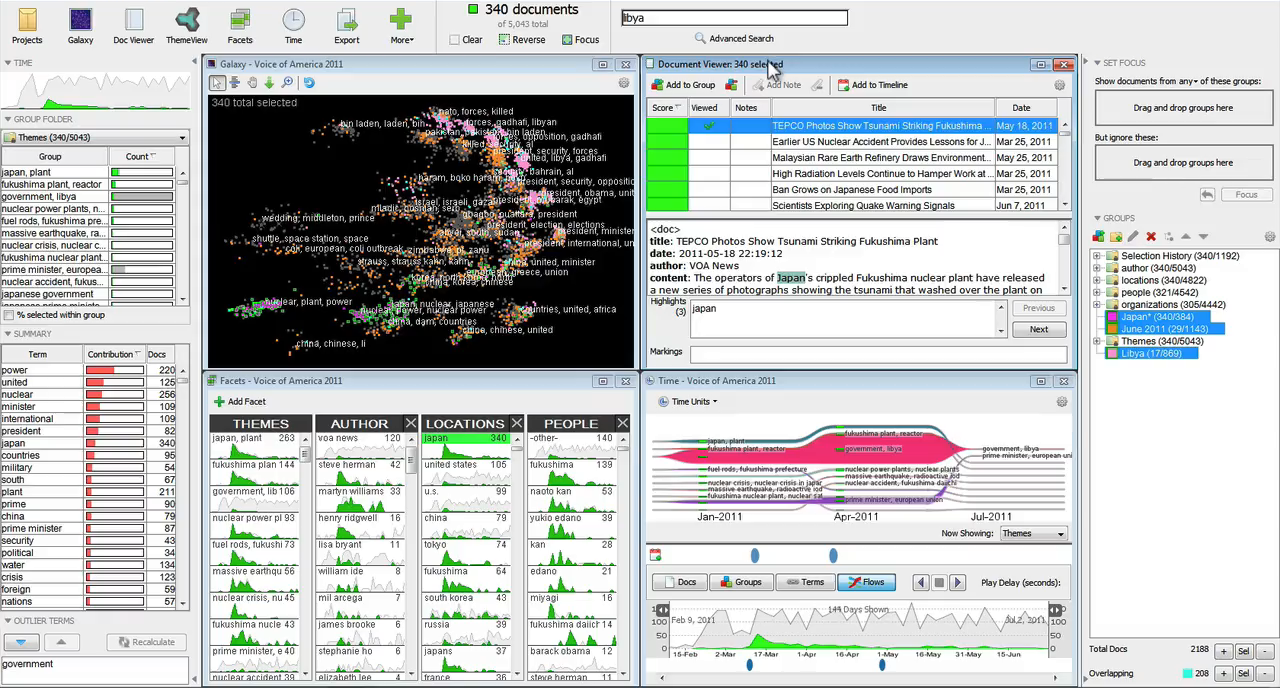}
  \caption{\label{fig:inspire} IN-SPIRE~\cite{Wise1999Visualizing}, a VA system for text corpora. IN-SPIRE combines computational metrics with interactive visualizations.}
  \vspace{-2em}
\end{figure*}

Visual analytic tools have been used to support information foraging by representing high-dimensional information, such as text, in an easily comprehensible two-dimensional view. In such views, the primary representation is one where information that is relatively closer to other information is more similar (a visualization method borrowed from cartography~\cite{Skupin2002Cartographic}). These applications allow users to find relevant information and gain new insights into topics or trends within the data. An early example of combining machine learning with visual analytics for analyzing text is a system called \textit{IN-SPIRE}~\cite{Wise1999Visualizing}. One of the views of the system, the \textit{Galaxy View} shown in Figure~\ref{fig:inspire}, displays documents clustered by similarity. Using dimension reduction techniques, this view encodes relative similarity as distance (documents near each other are more similar). The high-dimensional representation of the text documents is created by keyword extraction from each document (defining a dimension), and weightings on the keywords determined computationally using popular methods such as TF-IDF, etc.~\cite{Rose2010Automatic}.


Visual analytic tools have also been used to support synthesis by enabling users to externalize their insights during an investigation. In a spatial workspace where users can manually manipulate the location of information, users build spatial structures to capture their synthesis of the information over time - a process referred to as ``incremental formalism''~\cite{Shipman1999Formality,Shipman2001Emergent}. Andrews et al. found that intelligence analysts can make use of such spatial structures as a means to externalize insights during sensemaking, manually placing relevant documents in clusters on a large, high-resolution display~\cite{Andrews2010Space}. Additionally, they found that the large display workspace promoted a more spatially-oriented analysis. Tools, such as I2 Analyst's Notebook~\cite{i2}, Jigsaw's ``Tablet view''~\cite{Stasko2008Jigsaw}, nSpace2~\cite{Eccles2008Stories,Wright2006Sandbox}, Analyst's Workspace~\cite{Andrews2012Analysts}, and others have also found it helpful to provide users with a workspace where spatial representations of information can be manually organized. 

More recently, researchers have developed techniques such as Latent Semantic Analysis (LSA) for extracting and representing the contextual meaning of words~\cite{landauer1997solution}. LSA produces a concept space that could then be used for document classification and clustering. Also, probabilistic topic models have emerged as a powerful technique for finding semantically meaningful topics in an unstructured text collection~\cite{blei2009text}. Researchers from the knowledge discovery and visualization communities have developed tools and techniques to support visualization and exploration of large text corpora based on both LSA (e.g.,~\cite{dou2012leadline,crossno2009lsaview}) and topic models (e.g.,~\cite{iwata2008probabilistic,liu2009interactive,wei2010tiara,oesterling2010two}).

The Latent Dirichlet Allocation (LDA) model of Blei et al.~\cite{blei2003latent}, which represents documents as combinations of topics that are generated, in the unsupervised case, automatically has proved particularly useful when integrated in a visual analytics system. The LDA model postulates a latent topical structure in which each document is characterized as a distribution over topics and most prominent words for each topic are determined based on this distribution. Each topic is then described by a list of leading keywords in ranked order. When combined with VA techniques, LDA provides meaningful, usable topics in a variety of situations (e.g.,~\cite{griffiths2004finding,zhu2007storylines,dou2011paralleltopics}). Recent developments in the ML community provide ways to refine and improve topic models by integrating user feedback, e.g. moving words from one topic to another \cite{Hu:Boyd-Graber:Satinoff:Smith}.

There have been extensions of LDA-based techniques and other text analytics by investigating texts in the combination <\textit {topic, time, location, people}>. This permits the analysis of the ebb and flow of topics in time and according to location~\cite{dou2011paralleltopics,dou2012leadline,luo2012eventriver}. Time-sensitivity is revealed not only in topics but in keyword distributions~\cite{dou2012leadline}. Lately there has been work to add people and demographic analysis as well~\cite{dou2015demographicvis}. Combining topic, time, and location analysis leads to identification of events, defined as ``meaningful occurrences in space and time''~\cite{krstajic2011cloudlines,dou2012leadline,cho2016vairoma,luo2012eventriver}. Here the topic analysis can greatly help in pinpointing the meaning. In addition, combining topic modeling with named entity extraction methods, such as lingpipe~\cite{2008Alias}, can greatly enhance the time, location, and even people structure since these quantities can be automatically extracted from the text content~\cite{maceachren2011senseplace2,cho2016vairoma}.

\begin{figure*}
  \centering
  \includegraphics[width=0.8\linewidth]{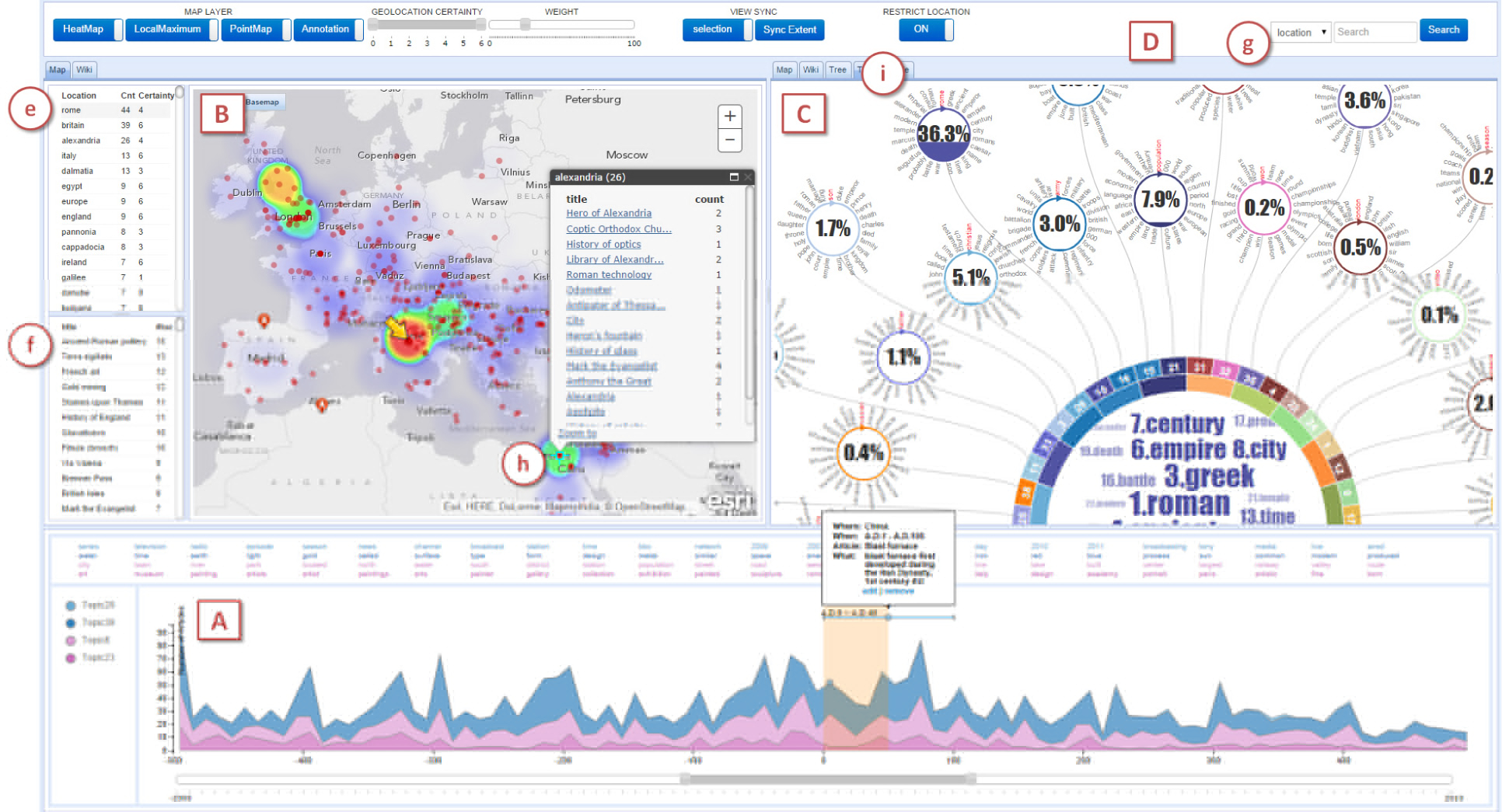}
  \caption{\label{fig:vairoma} Overview of VAiRoma Interface. The interface has three main views: Timeline view (A), Geographic view (B) and Topic view (C). A user-generated annotation is shown in the Timeline view.}
  \vspace{-2em}
\end{figure*}

At this point, it is worthwhile to describe a visual analytics system that combines all these characteristics. VAiRoma~\cite{cho2016vairoma} (shown in Figure~\ref{fig:vairoma}) creates a narrative that tells the whole 3,000 year history of Rome, the Empire, and the state of Italy derived from a collection of 189,000 Wikipedia articles. The articles are selected from the nearly 5M English language article collection in Wikipedia using a short list of keyword, but otherwise the initial topic modeling and named entity extraction are done automatically. The interface for VAiRoma is displayed in Figure~\ref{fig:vairoma}. The individual topics are depicted as color-coded streams in the timeline view (A). The circular topic view in (C) provides a compact way of depicting topics, the weights of their contributions for a given time range, and topic keywords. The navigable map view in (B) provides immediate updates of geographic distribution of articles (based on locating the geographic entities in the text) in terms of hotspots for a selected time range and topic. The window (f) lists article titles for selected geographic view, time range, and topic. In Figure~\ref{fig:vairoma}, one can clearly see event peaks for selected topics having to do with Roman government and military battles in the period from 500 BC to 500 AD. The interlinked windows in the interface plus key topics and event peaks permit a user to quickly peruse the main events in ancient Roman history, including the rise of Christianity and the Catholic church, trade with India and the Far East, and other events that one might not find in looking narrowly at, say, just the history of the Roman Empire. In this case, the user can focus from thousands of articles to a few hundred articles overall, which she can then quickly peruse. See the VAiRoma article for more details.

VAiRoma shows the power of the overall model depicted in Figure~\ref{fig:sacha_model}. Though it is not complete w.r.t. this model (no current VA system is), it provides an integrated approach to data handling, interactive visualization, ML (in this case topic modeling) combined with other techniques, and exploration and knowledge building techniques. It shows the power of an integrated approach. The approach is general and is now being applied to large, heterogeneous collections of climate change documents. In addition, full text journal article collections are being analyzed using extensions of the topic modeling and entity extraction methods. This shows that once <topic, time, location, people> features and event signatures can be extracted, analyses based on these analytics products can integrate a wide range of heterogeneous collections. 

\subsection{Multimedia Visual Analytics}
Visual analytic applications have also been developed to allow people to explore multimedia (i.e., images, video, audio). For example, iVisClassifier shows how facial expression features can be incrementally explored and classified by a combination of image feature-detection algorithms and user feedback~\cite{choo2010ivisclassifier}. Through interactively adding and removing images from classifiers, the model learns the facial expressions that are interesting (and similar) to the user. It combines analytic models such as feature extraction and classification with visual analytic approaches. \textit{MultiFacet} is another example of visually analyzing multimedia data~\cite{Henry2013MultiFacet}. MultiFacet presents facets of each data type to users as interactive filters. Thus, the process of interactively selecting attributes of different data types helps create groups of conceptually interesting and related information. 

As image and video data is often combined with text data (or textual metadata attached to the images or videos), fusing the feature space between these datatypes is an open challenge. Automated approaches are error-prone, and often require user intervention and guidance when semantic concepts and relationship need to maintained across data types~\cite{choo2012heterogeneous}. Similarly, an example of a much more specific application is given in \cite{BryanMysore2013} where the authors present a steering mechanism for source separation in a single monophonic recording. The user can annotate a standard time-frequency display to roughly define the different sources. Errors made by the algorithm can be annotated to improve further the separation.


\subsection{Streaming Data: Finance, Cyber Security, Social Media}
Streaming data is a growing area of interest for visual analytics. Data are no longer isolated and static, but instead are part of a sensor-laden ecosystem that senses and stores data at increasing frequencies. Thus, visual analytic systems that integrate machine learning models have great potential. Examples of domains that generate streaming data include the financial industry, cyber security, social media, and others. 

\begin{figure}[t]
  \centering
  \includegraphics[width=0.99\columnwidth]{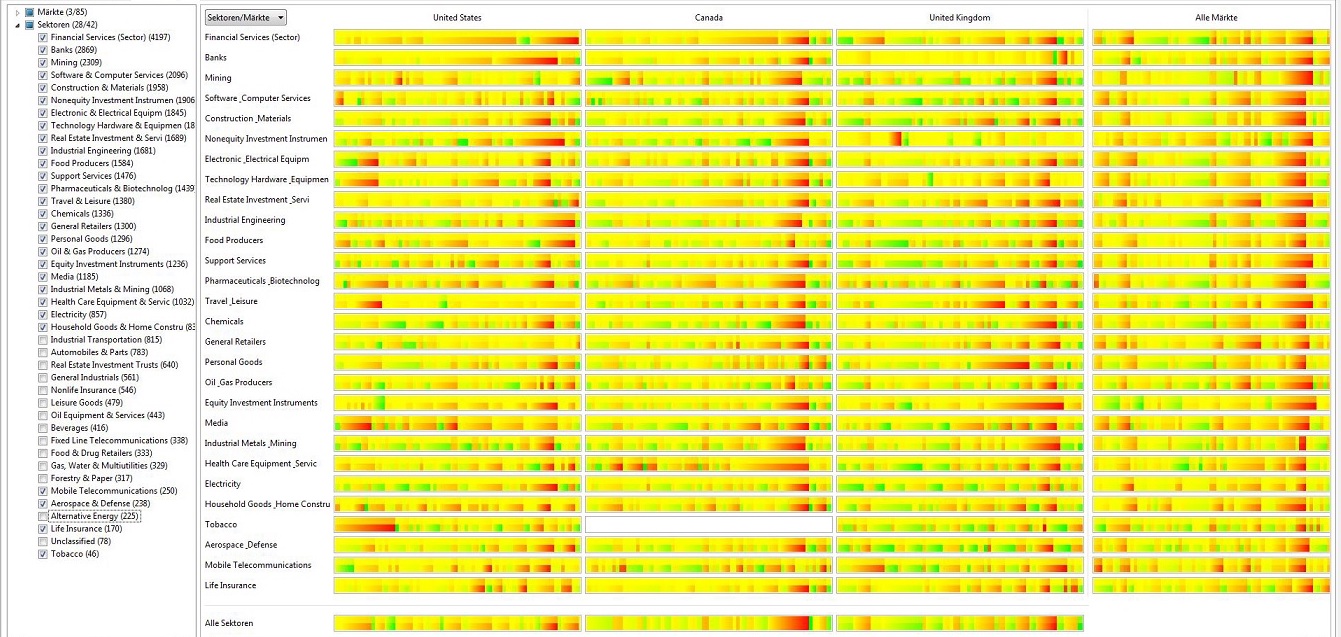} 
  \caption{\label{fig:financialViz} Aggregated visual representations and clustering have been used in supporting the real-time analysis of temporal sector-based market data~\cite{ziegler2010visual}.}
  \vspace{-2em}
\end{figure}

In finance, for example, \textit{FinVis} is a visual analytics system that helps people view and plan their personal finance portfolio~\cite{rudolph2009finvis}. The system incorporates uncertainty and risk models to compute metrics about a person's portfolio, and uses interactive visualizations to show these results to users. Similarly, Ziegler et al. presented a visual analytic system to help model a user's individual preferences for short, medium, and long-term stock performance~\cite{ziegler2008visual} and later extended their approach to real-time market data~\cite{ziegler2010visual}. Figure~\ref{fig:financialViz} is an example of how visualisations can provide an in-depth understanding of the groupings (clusterings) of financial time series. Here, financial market data for assets in 3 countries and 28 market sectors from 2006 and 2009 are depicted. The red bars indicate the crash of the stock market in 2008 and the visualisation enables the user to identify the overall changes but also notice subtle variations such as the lack of a response in some countries for particular sectors. 

Cyber security is a domain fraught with fast data streams and alerts. Examples of machine learning techniques often incorporated into systems that support this domain include sequence and pattern-based modeling, rule-based alerting, and others~\cite{Best20147}. People in charge of the safety and reliability of large networks analyze large amounts of streaming data and alerts throughout their day, thus the temporal component of making a decision from the analysis is emphasized. For example, Fisher et al. presented \textit{Event Browser}, a visual analytic system for analyzing and monitoring network events~\cite{fisher2012real-time}. Their work emphasizes how different tasks of the analyst have to happen at different time scales. That is, some tasks are ``real-time'', while others can be taken ``offline'' and performed for a longer duration of time. The persistent updating of new data into the offline tasks presents challenges. 

Social media data can also be analyzed using visual analytic systems. For example, \textit{Storylines}~\cite{zhu2007storylines} and \textit{EventRiver}~\cite{luo2012eventriver} are two examples of how visual analytic applications can help people understand the evolution of events, topics, and themes from news sources and social media feeds. In these systems, similar machine learning techniques are used as for text. However, the temporality of the data is more directly emphasized and taken into account.  

Lu et al.~\cite{Lu2014Integrating} showed how appropriate social media analysis could have predictive power, in their case predicting movie box office grosses from early word of mouth discussion on Twitter, YouTube, and IMDB. A dictionary-based sentiment analysis was used along with analytics from the R statistical computing environment and the Weka machine learning workbench. This permitted a choice of modeling in terms of multivariate regression, support vector machines, and neural networks. The paper promoted an integrated visual analytics approach where the interactive visualizations, based on D3, permitted users to investigate comments and sentiment, classify similar movies, and follow trends and identify features. The user could then improve a base line regression model based on trends and features identified in the visaulizations. Results of the use cases were positive with several of the non-expert participants being able to outperform experts in predicting opening weekend grosses for 4 films, according to the criteria set up by the authors. The paper has the usual limitation of supervised learning approaches in that a training dataset must first be collected and analyzed as a preliminary step, but it does successfully allow for improvement of the analytic model within the VA environment. Also, like many papers dealing with more complex analysis, it defines a process for best use of the system; this appears to be an important and effective approach for VA + ML systems.

Yeon et al.~\cite{yeon2016predictive} covered similar ground in their identification and analysis of interesting past abnormal events as a precursor for predicting future events. Here, as in Lu et al. and in other papers using ML, context and analytic power is obtained from combining multiple sources (in this case social media and news media). Yeon et al. identify contextual pattern in these past events, which permit them to make predictions for future events in similar contexts. An interactive interface involving spatio-temporal depiction of events plus identification of other features permits the choosing of interesting events and specification of their contexts. Trends for the unfolding of future events and possible unfolding story lines can then be created. The authors evaluated their VA system with three use cases.

\subsection{Biological Data}
Biology, and in particular, bio-informatics are fields that are increasingly becoming data-rich and the use of visualisation empowered analysis methods are proving highly useful and effective~\cite{gehlenborg2010visualization}. Although most computational analysis solutions only incorporate visualization as a communication medium and do not make use of interaction, there are a number of examples where  VA and ML approaches operate in integration. 
Within the context of epigenomic data analysis, Younesy et al.~\cite{younesy2013interactive} present how a number of ill-defined patterns and characteristics within the data can be identified and analysed through the help of interactive visualizations and integrated clustering modules. They demonstrate how user-defined constraints can be utilised to steer clustering algorithms where the results are compared visually. Grottel et al.~\cite{grottel2007} discuss how interactive visual representations can be instrumental in interpreting dynamic clusters within molecular simulations. In addition to these, interactive visualisations have been shown to support bi-cluster analysis~\cite{streit2014furby}. The authors utilize an interactive layout where fuzzy bi-clusters are investigated for multi-tissue type analysis. Biclustering is an algorithmic technique to solve for coordinated relationships computed from high-dimensional data representations~\cite{madeira2004biclustering}, and has been used in other domains, including text analysis~\cite{sun2014five,sun2016biset,Fiaux2013Bixplorer}.

In addition to the above methods where the focus is mainly on investigating clusters, there are also works where interactively specified high-dimensional data projections are utilised to characterize and compare different cancer subtypes~\cite{amir2013visne}. In their tool called viSNE, the authors demonstrate how user-driven, locally applied projections preserve particular relations and they argue that such methods are instrumental in interpreting any  multi-dimensional single-cell technology generated data.

\section{Embedding Steerable ML Algorithms into Visual Analytics}
\label{subsec:SteeringDR}

As discussed above at several points and categorized in Section ~\ref{sec:category}, one area of research that has been recently attracting much interest in the machine learning and data visualization communities is the development of interactive approaches binding visualizations to steerable ML algorithms. This goes beyond typical interactive ML methods in that it places interaction at the same level as visualization and ML, thus producing a powerful extension of visual analytics. As explained in \cite{Wijk05}, \cite{pike2009science}, interaction provides feedback in the visualization process, allowing the user to manipulate the parameters that define a visualization on the basis of the knowledge acquired in previous iterations. In particular, low latency interaction with large update rates of the visual display provides higher levels of user involvement in the analysis \cite{elmqvist2011fluid}, triggering low level attention and processing mechanisms (such as tracking moving items), where the user's senso-motor actions have immediate effects in the displayed information. Despite interaction mechanisms having extensively been discussed in the visualization literature \cite{Wijk05}, \cite{pike2009science}, the relationships between these parameters and the resulting visualization are in most cases of a simple nature, including changes of scale, displacements, brushing, etc., specially for low latency interaction.  As pointed out in \cite{Verleysen13}, hardly ever are complex interactions or transformations based on intelligent data analysis undertaken at this level. This fact is certainly surprising, especially considering that ML is a mature discipline and the power of today's hardware, as well as programming languages and libraries make it possible to use algorithms (or adapted versions of them) as intermediates between the user actions and the visualization, even at low latency levels.

The DR algorithms discussed in Section~\ref{sec:category}, which construct a mapping from a high dimensional input space onto a typically 2D or 3D visualization space, would be  particularly useful for extended VA approaches. To build such mappings, DR algorithms seek to preserve neighborhood relationships among the items in both spaces, resulting in representations that follow the so called ``spatialization principle'' (based on the cartographic principle where closeness $\approx$ similarity~\cite{Skupin2002Cartographic}). Placing similar items in close positions results in highly intuitive arrangements of items in a visual map that serves as a basis for developing insightful visualizations of high dimensional elements~\cite{Vesanto99,Kaski11,endert2013beyond}. Moreover, the connection that DR mappings make between something that can be ``seen'' and a high dimensional feature space suggests using the visual map as a canvas where classical interaction mechanisms (zoom, pan, brushing \& linking, etc.) can be used to explore high dimensional data. 

However, interaction can go far beyond this point by allowing the user to steer the DR algorithm through the visualization by direct modification of its parameters or by making transformations on the input data. As discussed in Section~\ref{sec:category}, this idea has been explicitly formulated in~\cite{Choo13} as \emph{iteration-level interactive visualization}, which aims at visualizing intermediate results at various iterations and letting the users interact with those results in real time. In a slightly more formal way, as shown in \cite{Diaz16}, an interactive DR algorithm --the argument can be extended to other ML algorithms-- can be considered as a dynamically evolving system, driven by a \emph{context} that includes the input data and the algorithm's parameters 
\begin{eqnarray}
\dot {\mathbf y} &=& \mathbf f(\mathbf y, \mathbf u), \qquad \mathbf v = \mathbf g(\mathbf y) \label{iDRmodel}
\end{eqnarray}
where $\mathbf y$ is the \textit{internal state} of the algorithm, $\mathbf v$ is the \textit{outcome} of the algorithm (e.g. a visualization), which depends on the internal state, and $\mathbf u = \{\mathbf x, \mathbf w\}$ is a \textit{context vector} that contains the \textit{input data} $\mathbf x$ and the \textit{algorithm parameters} $\mathbf w$. In a general framework, the user will steer the algorithm by manipulating $\mathbf w$ based on his/her knowledge acquired from the visualization $\mathbf v$. Under a fixed context $\mathbf u^0$ --i.e. no changes in the input data or the algorithm parameters--, the internal state $\mathbf y$ in model (\ref{iDRmodel}) will keep on changing until it reaches convergence to a steady state condition $\mathbf 0 = \mathbf f(\mathbf y^0, \mathbf u^0)$. Changes in the algorithms parameters $\mathbf w$ or in the input data $\mathbf x$ will make the internal state evolve to a new steady state condition $\mathbf 0 = \mathbf f(\mathbf y^1, \mathbf u^1)$, and hence result in a new visualization $\mathbf v^1$. For a continuous $\mathbf f(\cdot)$ --typically for non-convex algorithms, based on gradient descent approaches-- the representation $\mathbf v(t)$ will smoothly change, resulting in animated transitions that provide a continuous feedback to the user. Despite the fact that this behavior opens a broad spectrum of novel and advanced user interaction modes and applications, this is still a rather unexplored topic. 

Many possibilities may arise from this approach, all based on changes in different elements of the context vector $\mathbf u$:
\begin{itemize}
\item One fundamental subset of parameters that conveys a great deal of user insight are the \textit{input data metrics}, which can be expressed as a weight matrix $\Omega = (\omega_{rs})$ being $\|\mathbf a \|_{\Omega} = \sum_r\sum_s a_r \omega_{rs} a_s$, whose parameters are included in $\mathbf w$. Prior knowledge on the \textit{relevance of features} can be easily considered allowing user-driven modifications in the diagonal elements of $\omega_{ii} \subset \mathbf w$.  An example related to this idea is the iPCA \cite{Jeong09}, an interactive tool that visualizes the results of PCA analysis using multiple coordinated views and a rich set of user interactions, including modification of dimension contributions. A similar idea on the stochastic neighbor embedding algorithm (SNE) was also proposed in \cite{Diaz14}. 

\item The user might also have insight on the \textit{similarities between items}. In \cite{brown2012dis}, a system called dis-function was developed, featuring DR visualization that allows the user to modify the distance matrix $D_{ij}=\|\mathbf a_i - \mathbf a_j\|_\Omega$ between items $i$, $j$, by moving points in the visualization based on his/her understanding of their similarity, and see new results after a recomputation of the projections with the new metrics. 
\item Also, \textit{prior knowledge on class information} can be inserted by the user, suggesting techniques to increase the similarity of items belonging to the same class. In \cite{Perez15} a method is proposed to allow the user to include prior class knowledge in the DR projections by extending the original dataset with transformations of the original feature space based on his existing class knowledge. 
\item Finally, the input data $\mathbf x$ in  model (\ref{iDRmodel}) may change with time ($\mathbf x = \mathbf x(t)$), suggesting the use of iDR on streaming data to provide live visualizations $\mathbf v(t)$ that convey \textit{time varying information}; in this case, user interaction is possible through timeline sliders, making it possible to explore how input data items and their relationships evolve in time by moving back and forth in time.
\end{itemize}

These cases imply a substantially more advanced kind of feedback to the user than traditional interaction mechanisms. Placing these capabilities in a visual analytics framework greatly empowers them. As described in Figures 2 and 3, such a framework supports analytic reasoning, the discovery of much deeper insights, and the creation of actionable knowledge. The mere fact of being part of sensemaking and knowledge feedback loops (a virtuous cycle) suggests that there is huge potential and a broad spectrum of possibilities in the integration of ML algorithms discussed in this paper, where even the simplest ones may have multiplicative effects. For certain types of analysis, such as following animated transitions, this sort of interaction mechanism must be achieved in a fluid manner, with low latencies and fast update rates. However, this is not necessarily required for all knowledge generation and synthesis activities, as discussed next.

\paragraph*{Levels of Interactive Response}
A long-recognized upper threshold for latency in WIMP and mobile interfaces is 0.1 second. Faced with higher latencies, users start to lose the connection between their actions and the visual response, commit more typing or selection errors, and become frustrated~\cite{hoober2011designing}. This limit has also been discussed as an upper threshold for coherent animations (though completely smooth animations would require a lower latency) and for a range of interactions in immersive VR. However, the detailed effects of particular latency thresholds depend on the task. For embedded analytics tools in VA systems, such as steerable ML methods, it is useful to define a wider range of interactive responses~\cite{Ribarsky2016Human}:
\begin{itemize}
\item \textit{Real-time Regime}: < 0.1 second. Interactions such as moving a time slider to control an animation of time-dependent behavior or changing the weighting factors of leading dimensions in an interactive PCA tool~\cite{Jeong09} to reveal changes in the projected surface fall into this regime. Such interactions can be employed for rapid exploration and spotting of trends.
\item \textit{Direct Manipulation Regime}: 0.1 to 2-3 seconds. Analytic reasoning tends to involve more complicated interlinking of rich visualizations with ML methods. For example, the VAiRoma geographic window shows multiple hierarchical hotspot clusters (Figure~\ref{fig:vairoma}) when a time range and topic are selected, but there is a delay of 2-3 seconds before the result is displayed. The user must peruse this distribution and its areas of concentration, which can take several seconds or more. During interface evaluation the delay was not noted and does not seem to hinder the user's reasoning process~\cite{cho2016vairoma}, perhaps because the user is thinking about the selection when it is made, and what it may mean, which then flows into her reasoning process once the result appears. The same seems to be true when the user makes a selection of a geographic region or a topic and experiences a similar delay until updates in the timeline or other linked windows appear.
\item \textit{Batch Regime}: 10 seconds or more. Here the cognitive flow of human reasoning is interrupted. To minimize effects of this interruption, the best analytics at this level of response might be those that launch a new reasoning direction (e.g., recalculation of textual topics based on a revised set of keywords).
\end{itemize}
These levels of response are related to performance timings from enactive cognition~\cite{gray2006soft}, suggesting that this model can be applied here. An important conclusion of this discussion is that it is not necessary to have real-time response for certain interactive ML algorithms; delays up to 2-3 seconds and perhaps more might be digestible by the user. This could substantially reduce the burden of interactive response for ML algorithms. Of course, further user studies of these algorithms in action should be carried out.

\section{Open Challenges and Opportunities for ML and VA}
\label{sec:challenges}
Collaboration between ML and VA can benefit and drive innovation in both disciplines. Advances in ML can be used by VA researchers to create more advanced applications for data analysis. This includes the optimization of currently integrated techniques, but also the discovery of additional techniques that fit into the broad range of analytic tasks covered by visual analytic applications~\cite{Amar2005Low,lee2006task}. Similarly, as advances are made in VA applications, the user requirements and needs can drive new ML algorithms and techniques. 

Below, we list a collection of current challenge and opportunities at the intersection of ML and VA.

\subsection{Creating and Training Models from User Interaction Data}

ML models are typically built and modified based on ample training data that contain positive and negative ground truth examples. While many domains and tasks can be solved with ample training data, there exist scenarios, as discussed in this paper, where not enough training data is available. For these cases, it becomes important to incorporate user feedback into the computation in order to guide and parametrize the computational model being used. This raises the challenges of \textit{how to incorporate user feedback into computation in an effective and expressive, yet usable manner?}

The concept of interactive machine learning has taken into account user feedback to steer and train these models. For example, users can provide positive or negative feedback to give support for or against suggestions or classifications made by the model. The models adjust over time based on this input.

However, there is the ability to look beyond labeling, or confirming and refuting suggestions as way to incorporate user feedback~\cite{Endert2015Semantic} - what about the remaining user interaction that people perform during visual data exploration? User interaction logs contain rich information about the process and interests of the user. Examples of the kinds of inferences that can be made from the user interaction logs are shown in more detail earlier in the report. Thus, the opportunity exists for ML techniques to leverage the real-time user interaction data generated from the analysts using the system to steer the computation. 

Systems that take into account a broader set of user interactions enable people more expressivity in conveying their mental model, preferences, and subject matter expertise. Further, taking into account the broader set of user interaction allows users of the system to stay more engaged in the act of visual data exploration, as opposed to actively training the model and system. 

\begin{figure}
  \centering
  \includegraphics[width=0.6\columnwidth]{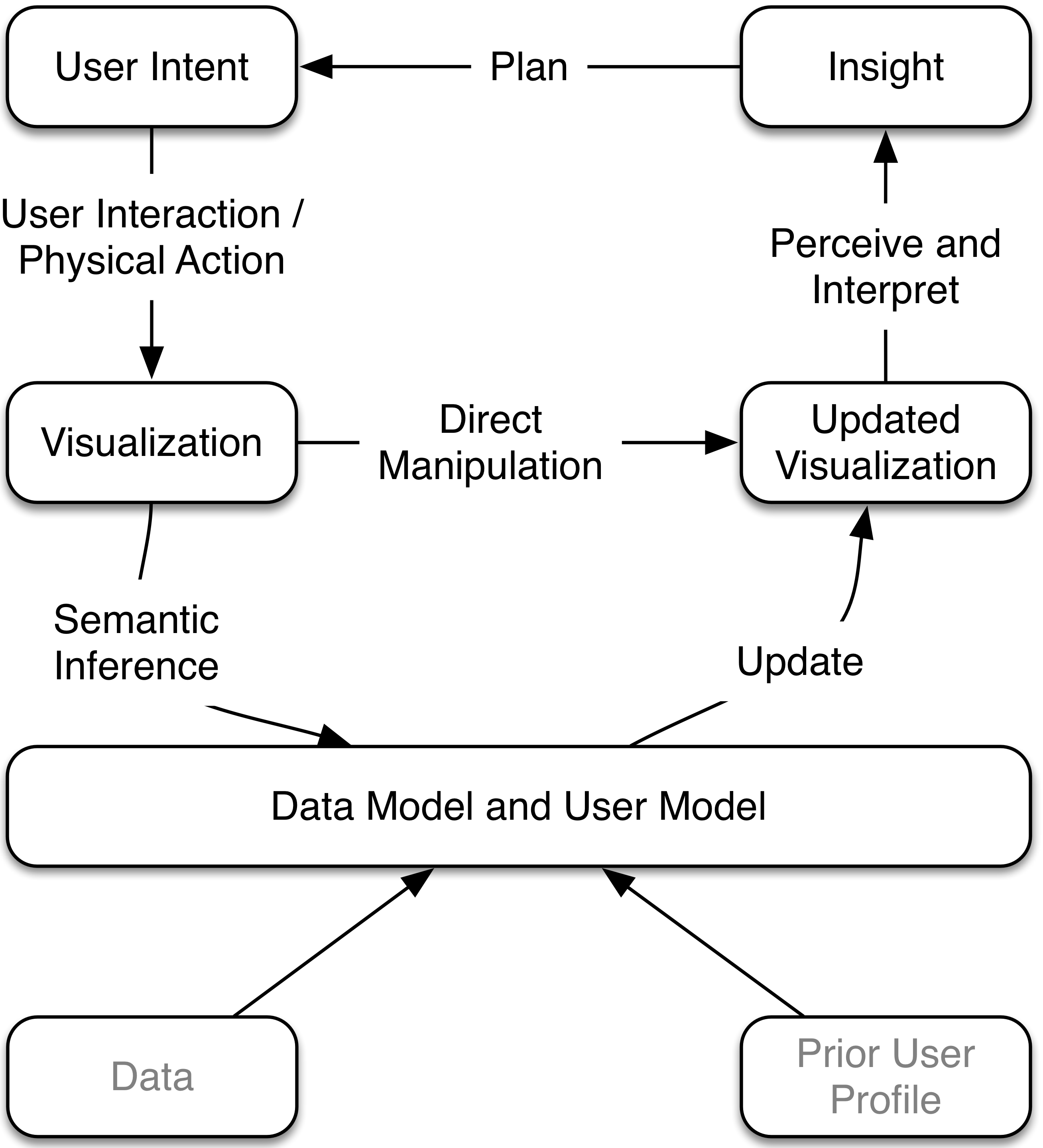}
  \caption{\label{fig:si_diagram} A model from~\cite{Endert2015Semantic} showing how multiple types of user input can be used to steer machine learning models in VA.}
  \vspace{-2em}
\end{figure}

Figure~\ref{fig:si_diagram} shows a model for how multiple types of user input can be incorporated into the machine learning models driving visual analytic techniques. As is shown in this model, two broad types of models can be created from user interaction: Data models and User Models. In general, data models refer to weighted data items and attributes. These can be weighted computationally, or via user feedback. Further, these weights can be computed based on inferences on the user interaction (i.e., to approximate user interest of focus). User models typically refer to computational approximations of the state of the user (e.g., cognitive load, personality traits~\cite{Brown2014Finding}, etc.)

In addition to steering existing models (such as dimension reduction models, topic models, etc.), such user feedback can indicate the need for novel models to be created. By focusing on the user interaction, new discoveries can be made about the processes and analytic tasks of people during data analysis. This continued study, or \textit{science of}, interaction~\cite{pike2009science} can lead to advances in the machine learning community in the way of new algorithms or techniques that model analytic tasks or processes of people. 

\subsection{Balancing Human and Machine Effort, Responsibility, and Tasks}

For mixed-initiative systems, it is a common notion that there exists a balance of effort between the user and the machine~\cite{Horvitz1999Principles}. This effort can be divided by decomposing the larger task into sub-tasks that are either better suited to the person, or more quickly performed by the system. Similarly, these tasks often break down into being more well-defined and quantitative (i.e., solved by computation), or subjective and less formally defined (and thus needing input from the user). For example, a mixed-initiative visual analytic system for grouping and clustering can take into account the exemplar data items that are grouped by the user, generate a data model from those examples, and organize the remaining data points~\cite{Drucker2011Helping}.

However, there remains the need for generalizable empirical evidence to inform researchers about how to balance this effort between the user and the machine. It is not clear the extent to which tasks should be divided, or co-completed. Typical data analysis sessions involve many user tasks and sub-tasks~\cite{Amar2005Low}, and dividing the effort of these tasks between the user and the system is challenging. 

It is also unclear exactly how to measure the amount of effort expended by both the user and the system. For example, in a visual analytic system that helps people cluster documents, Endert et al. used a measure of how many documents were moved and grouped by the user and how many were automatically grouped by the system~\cite{Endert2012Semantic-vast}. However, there exist opportunities to consider additional metrics for the balance of effort in mixed-initiative systems that can drive the possibility of novel evaluations of effectiveness.

\subsection{Complex Computation Systems can lead to Automation Surprise}

By coupling machine learning with visual analytics systems, we can develop complex systems made up of many inter-related and inter-dependent ``black boxes'' of automated components for data analysis, knowledge discovery and extraction. Complex systems will typically comprise many instances of known and hidden inter-dependencies between components and yield outputs that are emergent where the interactions among agents and individual units may be deterministic. The global behaviour of the system as a whole may conform with rules that are only sometimes deducible from knowledge of the interactions and topology of the system. This makes it difficult to know exactly which inputs contribute to an observed output, and the extent of each factor's contributions~\cite{satinover2011taming,ormandWhat}.  Sarter and Woods~\cite{sarter1997automation} observed that interactions between these tightly coupled automated ``black boxes'' can create consequences and automation surprises that arise from a lack of awareness of system state and the state of the world. This creates potential for error, complacency from trusting the technology, placing new demands on attention, coordination and workload.

At the risk of saying the obvious, an approach proposed by Norman~\cite{norman1986cognitive} to address some of the problems of controlling complex systems is based on observability and feedback. They are crucial for figuring out how a system works, and they help us affirm the mental models that drive our thinking and analysis of a problem or a device. Poor observability of automated advanced intelligent processes makes it difficult to evaluate if outcomes from the automated computations are within the bounds of normal or acceptable behavior, or whether our instructions to the system were correctly executed or what else was included in the execution that was not intended. Good mapping between designed action and desired action helps us anticipate and learn how to interact with the system. Good mapping also helps us see the connection between what the system was instructed to do, and the outcome of carrying out that instruction. 

One of the major challenges then, is for visual analytics designers to create designs that ``... facilitate the discovery  of meaningfulness of the situation ... not as a property of the mind, but rather as a property of the situation or functional problems that operators are trying to solve ... [by] developing representations that specify the meaningful properties of a work domain ... so that operators can discover these meaningful properties and can guide their actions appropriately''~\cite{bennett2011display}.

To create such a design, there is a need to have a conception of the analytical thinking and reasoning process that extends beyond the information handling and manipulation aspects that are frequently described. A focus group study with 20 intelligence analysts~\cite{wong2012black}, think-aloud studies with 6 analysts performing a simulated intelligence task~\cite{rooney2014invisque}, and think-aloud studies with 6 librarians carrying out a surrogate task of creating explanations from a literature review task~\cite{kodagoda2013using} provide insight into this analytical thinking and reasoning process. The results of these studies indicate that analyst make use of the various inference making strategies described in Section 2.1 - induction, deduction and adduction - depending upon what data they have, the rules for interpreting the data, and premise they are starting with and the conclusions they would make or would like to make. Furthermore, very often they would test the validity of the propositions they arrive at by practicing critical thinking - where they attempt to assess the quality and validity of their thinking and the data they use, the criteria they use for forming judgments, and so forth. In fact, critical thinking is so important that many intelligence analysis training schools have introduced it into their training. 

\begin{figure}
  \centering
  \includegraphics[width=0.99\columnwidth]{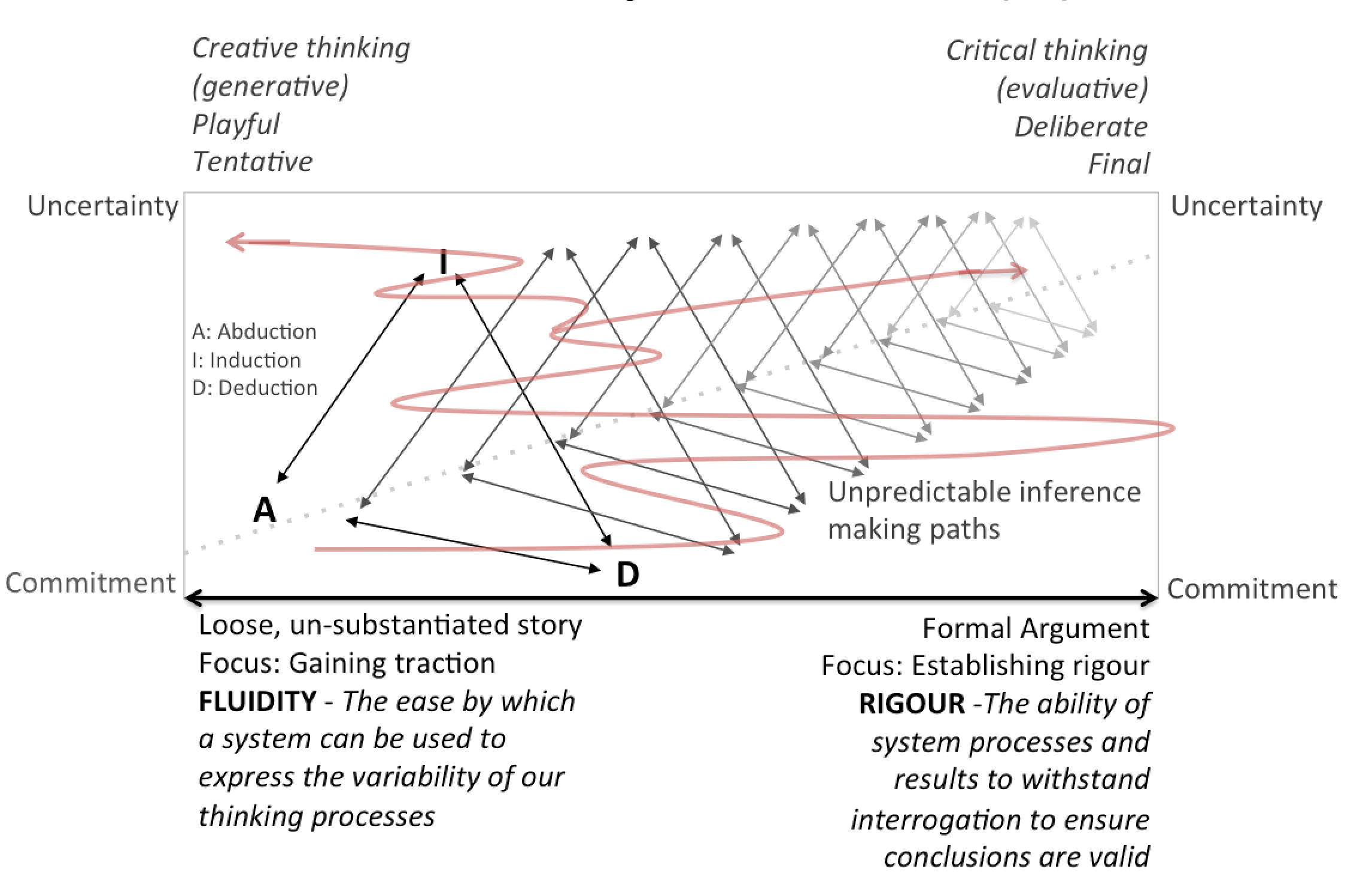}
  \caption{\label{fig:thinking_terrain} Characterizing the thinking terrain of analysts~\cite{wong2014analysts}.}
  \vspace{-2em}
\end{figure} 

One thing else that is observed to happen alongside all of this is somewhat more subtle: Analysts are constantly trying to explain the situation, sometimes re-constructing the situation from pieces of data and from inferential claims; and then carrying out searches or further analysis to find necessary data back the claims. This process of explanation is crucial to making sense and how it is used to link data, context and inferences. It often starts off as a highly tentative explanation that is based on very weak data or hunches. The analyst then explores this possibility, making conjectures, suppositions and inferential claims, from which they then connect with further data (testing their relevance and significance), elaborate, question, and often reframe and discard, their ideas, and eventually building up the story so that it eventually becomes robust enough to withstand interrogation.  

We see a progression - not necessarily in a linear manner - where explanations reflect tentative, creative and playful, and generative thinking, and then transitions towards thinking strategies that are more critical, evaluative, deliberate and final (see Figure~\ref{fig:thinking_terrain} for an illustration depicting this discussion).  One can assume a continuum where at one end we have a tentative explanation we call a ``loose story'' that accounts for the data, and at the other end the loose story has evolved into a strong and more formal argument such that it is rigorous and able to withstand interrogation, say, in a court of law.

At the ``formal argument'' end of the continuum, there is much lower uncertainty. The analyst is more definite about what the data and their relationships mean, and very likely has become more committed to a particular path of investigation. At this end, the emphasis is on verifying that the data used to construct the conclusions, the claims being made based on the data, and the conclusions themselves, are valid.

The combined machine learning and visual analytics tools to be built should fluidly link the generative, creative, playful and tentative exploration activities that encourage the exploration of alternatives, appreciation of the context, and the avoidance of pre-mature commitment, with the more evaluative, critical inquiry that leads to a deliberate, final and rigorous explanation.  This is the notion of the design principle of fluidity and rigour.

\subsection{Visualizing Intermediate Results and Computational Process}
Many kinds of ML algorithms undergo a continuous convergence process towards the final solution. In general, only this final solution is rendered into a visualization, which may incorporate classical interaction mechanisms (zoom, pan, brushing, focus\&context, etc.). This convergence is often done within a fixed context, that includes the training set, the algorithm parameters and the cost function. These elements often convey a large amount of insight for the user, but since they remain fixed during convergence users are deprived of the benefits of interaction. What if the user could steer these fixed elements ``during''  convergence?. 

A promising topic, involving innovation by both VA and ML communities, is rendering visualizations of the intermediate results during convergence, allowing the user to tune\/steer the ML algorithms by changing these elements. Designing \textit{ad hoc} ML algorithms with this approach in mind that pave the way for new and useful kinds of interaction mechanisms opens new and exciting research paths. There has been some prior work on this topic. For example, Stolper et al. developed a system for \textit{progressive visual analytics}, where intermediate results of a sequence-mining algorithm running on medical treatment events can be shown to clinicians~\cite{stolper2014progressive}. Their work gave analysts the ability to see broader results sooner to help decide if the entire computation needed to be executed. Similarly, systems to show partial query results of large datasets~\cite{Fisher2012Trust} and partial dimension reduction and clustering results~\cite{turkay2017Progressive} have been recently developed.. These works raise important questions about the tradeoff between accuracy and execution time of these algorithms, and also about how to incorporate user feedback into computation during runtime. 

\subsection{Enhancing Trust and Interpretability}

A key element of the visualization approach is its ability to generate trust in the user. Unlike pure machine learning techniques, in a data visualization the user ``sees'' the data and information as a part of the analysis. When the visualization is interactive, the user will be part of the loop and involved in driving the visualization. In such a context, the development of a mental model goes hand in hand with the visualization, as everything is part of the process. This tight involvement of the user in the development of the visualization based on the results of previous iterations, along with the highly visual component of human thinking, can make this approach generate a great amount of trust in the user. However, such ``trust'' can have different meanings at different levels of cognition. An apparently trustable result at an intuitive level can arouse suspicions at a higher cognitive level, demanding methods for statistical confirmation of the results. On a broad view, two different levels can be identified:

\begin{enumerate}
\item A ``qualitative level'', that would make heavy use of perception visualization principles along with interaction mechanisms to present data in an intuitive way. The communication in both senses (from and to the interface) will typically seek to: a) adapt to individual's perception mechanism so that the information throughput and knowledge increment on the user is maximized; and b) in a higher level, to adapt to the human cognitive process so that data and information is presented in a way that is intuitive to the user. The means to carry out this approach would rely on classical visualization methods (adequate use of visual encodings and spatial layouts) and on interaction techniques, including brushing, linking, coordinated views, animated transitions, etc., but also in much more powerful approaches such as user-driven steering of ML algorithms (such as DR, clustering, etc.) resulting in the reconfiguration of the visualization on the basis of changes in the context such as time varying data or changes in the user focus on different types of analysis. 

\item A ``quantitative level'' is, however, needed to provide sound statistical validation of the former visualization results. Taken in an isolated way, this level would lack insight. However, its outcomes are supposed to be trustworthy so the user can consider them as definite validations. Quantitative approaches --mainly belonging to the realm of ML-- are in essence deterministic, which makes them less prone to human errors and  reproducible. This helps to standardize decisions and provides congruence, accurateness, uniformity and coherence in the results. 

However, quantitative approaches tend to avoid the need for user intervention by trying to automate the process. In general they do not look for human feedback but undertake as many human tasks as possible in the process, automating it to the maximum possible extent, aiming to avoid any kind of human subjectivity and seeking rigor (statistical, mathematical). But many problems in real life are built on sparse bits of knowledge coming from diverse domains. Moreover, such knowledge is often made of vague or imprecise mental models. Purely quantitative approaches cannot operate with such small, diverse and ``fuzzy'' bricks; they need solid foundations to be operative.

\end{enumerate}

The previous division is only conceptual. Both approaches can (and should) be combined. For instance, a statistical validation of one or more facts can be displayed on top of the qualitative visualization by making use of visual encodings and text labels. We encourage visual analytics designers to seek efficient combinations between qualitative and quantitative approaches, looking for concurrent visualization of actual problem data and sophisticated computed features, both coexisting in the same representation. The mere fact of representing statistical validations sharing the same layout and structure as the original data in a same visualization allows the user to internalize that quantitative information allowing her to connect it to its domain knowledge, with an unquestionable positive effect on trust and confidence in the results.

\subsection{Beyond Current Methods}
Currently, many of the applications of machine learning in visual analytics relate to dimensionality reduction. In addition, as discussed in Section 4, there are a different sort of ML methods based on Bayesian inferencing and including topic modeling and textual analytics approaches. These are becoming more prominent. While these applications are undeniably an important use of machine learning, we contend that consideration of the role of the user opens up several new fields of study where machine learning can play an important role.  First amongst these is the role of machine learning in creating a computational model for the user's analytical process.  This complements cognitive task analysis and aims to model how domain expert users use visual analytics to tackle important tasks, and how they reason about the problem.  This will enable better system design to support expert strategies and provide support to less-trained users.

Every user interaction has two primary functions: i) to communicate a direct \emph{explicit} intent from the user to the analytical system and receive an appropriate response (e.g. if the user requests a zoom into a particular area, the system should create that zoomed-in visual display), and ii) to carry out an indirect \emph{implicit} piece of analytical reasoning.

The point is that every user choice in the visual analytics frame is equivalent to a statistical choice in the mathematical frame: we need users to make appropriate choices that do not invalidate the (implied) statistical analysis that they are carrying out. Motivated by the analysis of how users carry out visual analytics, particularly the concepts of sense-making and knowledge generation, the first step to understanding the details of this process is to compile a complete log of users' analytical process and the information that they record.  This is the base dataset that can be used for traceability, responsibility and provenance: providing an argued case for others (such as collaborators or managers) to critique and use to make decisions.  However, beyond this use, the database is also a resource to mine in order to clarify the decisions that are made in the course of visual analytics, leading to the potential to develop adaptable interfaces and a greater depth of understanding of users' mental models, which can then be used to guide other, perhaps less skilled or experienced, users.  

It would not be feasible (nor practically useful) to track every single change in a visualisation.  It is essential that the process involves minimal interruption to cognitive flow (so as to avoid damaging the very process we are trying to understand).  However, it would be helpful to prompt the user for feedback (preferably in visual ways), in the form of annotations, at certain key points of the analysis.  We propose using machine learning (e.g. to look for breakpoints in the way information is displayed) as cues for these prompts.  The process model can also learn from user interaction (with appropriate additional guidance).  For example, if the user \lq undoes\rq\ a particular action, it could mean \lq\lq I don\rq t want this: my choice was wrong\rq\rq\ or \lq\lq The visualisation is useful, but it is a dead end and I need to back-track\rq\rq.  Other simple user interactions that can connect to reasoning processes include brushing data points (which corresponds to selecting and labelling a subset of data) and linking (which corresponds to hypothesising correlations between variables and data points).

As a complement to this database of successful analytic practice, what many users need is a way of avoiding bad practice (or errors).  A catalogue of \lq typical\rq errors that is searchable (using case-based reasoning tools) could be crowd-sourced from teachers (and their students!) or training courses. 

How can machine learning aid the understanding of user processes?  At the simplest level, user interactions are a linear sequence of actions: discovering the underlying sequence and the transitions between items is relatively straight-forward, since a Markov (or hidden Markov) model can easily be trained to uncover this structure.  However, an unstructured and unannotated sequential list does not contain enough structure to infer the analytical process. Firstly, we need to understand the reasons why a user has made choices (which requires annotations). Secondly, it is clear that the analytical process is not a simple sequence of logical choices leading inexorably to a goal.  Instead, the process involves exploratory analysis -- trying a range of options and assessing which is the most successful -- and back-tracking when results show that a particular line of inquiry is fruitless.  These transform what is, in terms of a graphical model, a one-dimensional structure, into a tree or directed acyclic graph.  

The theory of Bayesian belief networks (BBNs) is relevant here.  There are two aspects of the model that can be learned: the \emph{conditional probability tables} (CPTs) for the links from all the parents of a particular node; and the \emph{structure} of the network (the presence or absence of directed links) which represents the conditional (in)dependence of variables.  Learning the CPTs for a given network structure is straightforward: with suitably chosen Bayesian priors (a Dirichlet distribution), it is a matter of counting co-occurrences of value pairs in a dataset~\cite{spiegelhalter1993bayesian}.  Learning the structure of a BBN is much more complex: in fact, the general case is NP-hard~\cite{chickering1996learning}.  Some special cases (such as trees) are tractable, but in this domain it is preferable to fix the structure based on our understanding of the users' analytical process. Models for this process, such as CRISP-DM~\cite{wirth2000crisp} (used in data mining) or those drawn from the infovis community (such as the semantic interaction pipeline), are currently rather high-level, and a more detailed task analysis is necessary before the requisite level of detail for a full computational model can be achieved.  

Once a computational user model for the analytic process is established, there are a number of other ways machine learning and visual analytics can be brought into dialogue.
\begin{enumerate}
\item Semi-automated report generation.  Machine learning can be used to infer links and relations between concepts, data, and analytical results, while frequentist or Bayesian statistical analysis can be used to attach a statistical significance to each finding.  This could be presented to the user as a checklist of automatically discovered analytical findings (or hints) that the user can accept or reject.
\item Annotations can be categorised using automated topic analysis (for example by Natural Language Processing that uses probabilistic graphical models~\cite{lin2010comparative}).  The value of this is to link annotations and find common approaches to tasks.
\item Model-based layout.  The goal is to provide a semi-automated way of modifying the layout of visual information.  One aspect of this is related to the steerable DR discussed in Section 5.  This can be extended to learning the criteria that analysts use: for example, how the user selects principal components.
\item Extreme value theory~\cite{de2007extreme} to identify low-frequency (but potentially high-value) data points or variables.  Recent research in this area supports the automated identification of outliers even in the multivariate case.
\item Integrated prior knowledge and data.  Often the expert user will have a great deal of prior cognitive knowledge embodied in a computational model of a physical system (e.g. geochemists supporting hydrocarbon exploration; meteorologists).  Machine learning can be used to generate an \emph{emulator}, a technique for model reduction that reduces the exceptionally high computational burden imposed by many physical models, while retaining the key features of the original model and allowing much greater user interaction for tasks such as sensitivity analysis and control~\cite{conti2010bayesian}.
\end{enumerate}

It is clear from the discussion throughout this paper that there are barriers to the closer integration of machine learning and visual analytics. One of the main technical barriers is that the current software tools are strongly divided between the research communities. Visualization tools are strong at close control over the form and layout of information, and user interaction: Some tend to be written as bespoke integrated tools, such as Tableau (http://www.tableau.com), Orange (orange.biolab.si) and JMP (www.jpm.com). On the other hand, the most advanced machine-learning tools are often written as libraries in numerical or statistical languages (such as Matlab, e.g. [Nab02] and R), as well, as in high level general purpose languages, like Java (with Weka, a widely used collection of ML algorithms for data mining tasks, or the Stanford NLP tools with advanced ML algorithms for natural language processing) or Python (with powerful and widely adopted data analysis and ML libraries like scipy, scikit-learn, pandas, etc.); all of them focus on supporting the (often) challenging task of learning complex models from data but provide limited graphical display and interaction. The best solution to this problem, short of reimplementing large toolkits in other languages is to take a client-server approach: a backend server running a good mathematical package for the machine-learning components complemented by web services and html+js clients, able to take advantage of the huge and growing spectrum of javascript libraries and frameworks (such as d3js) to provide interactive information visualisation.

\section{Conclusions}
This paper provides a comprehensive survey of machine learning methods, and visual analytics systems that effectively integrate machine learning. Based on this survey, we present a set of opportunities that offer a rich set of ideas to further the integration between these two scientific areas. Among these are formalizing and establishing steerable ML, generally providing coupled interaction and visualization methods that offer substantially more advanced user feedback. There is the opportunity to better determine how tasks should be divided between humans and machines, perhaps in a dynamic manner, including determining metrics for a balance of effort between these two components. The paper shows how recent models and frameworks could be used to develop considerably more powerful visual analytic systems with integrated machine learning. The summary and discussion presented in this paper seeks to excite and challenge researchers from the two disciplines to work together to tackle the challenges raised, ultimately creating more impactful systems to help people gain insight into data.

\section{Acknowledgments}
The authors would like to acknowledge that much of the content and inspiration for this paper originated during a Dagstuhl Seminar titled, ``Bridging Machine Learning with Information Visualization (15101)'' \cite{keim_et_al:DR:2015:5266}. In addition, funding for the authors was provided in part by the Analysis in Motion Initiative at PNNL, Spanish Ministry of Economy \& Competitivity and FEDER funds, under grant DPI2015-69891-C2-2-R 

\section{Author Bios}
\noindent\textbf{Alex Endert} is an Assistant Professor in the School of Interactive Computing at Georgia Tech, where he directs the Visual Analytics Lab. In 2013, his work on Semantic Interaction was awarded the IEEE VGTC VPG Pioneers Group Doctoral Dissertation Award, and the Virginia Tech CS Best Dissertation Award.

\noindent\textbf{William Ribarsky} is the Bank of America Endowed Chair and Director of the Charlotte Visualization Center at UNC Charlotte. He is one of the founders of the field of visual analytics and was Chair of the IEEE Visualization Analytics Science and Technology (VAST) Steering Committee until October, 2015. 

\noindent\textbf{Cagatay Turkay} is a Lecturer (Assistant Prof.) at the Department of Computer Science at City, University of London. He carries out his research at the giCentre and develops methods where interactive visualisations and computational tools are used in tandem for informed analysis processes. 

\noindent\textbf{William Wong} is a professor of Human-Computer Interaction at Middlesex University, London. He is the head of the Interaction Design Centre. His research interest include investigating the problems of visual analytics in sense-making domains with high information density, such as intelligence analysis, financial systemic risk analysis, and low literacy users.  

\noindent\textbf{Ian Nabney} is a Professor at Aston University.  He is Director of the System Analytics Research Institute.  His research interests are in machine learning, particularly in data visualisation, time series, and Bayesian methods.  His application interests are broad, and include condition monitoring, biomedical engineering, and urban science.  

\noindent\textbf{Ignacio D\'iaz Blanco} is an associate professor at the Department of Electrical Engineering at the University of Oviedo. He researches in intelligent data analysis, data visualization, control and signal processing to understand, diagnose and optimize processes and complex systems.

\noindent\textbf{Fabrice Rossi} is a Professor at Paris 1 Panth\'eon Sorbonne University. He is a member of the SAMM research group and the head of the statistical learning team of this group. His research interests include machine learning and data analysis. He is particularly interested in interpretable systems.

\bibliographystyle{eg-alpha-doi}

\bibliography{refs}

\end{document}